\documentclass[10pt,twocolumn]{article}
\usepackage{simpleConference}
\usepackage{times}
\usepackage{graphicx}
\usepackage{amssymb}
\usepackage[hyphens]{url}
\usepackage{hyperref}
\usepackage[numbers]{natbib}
\usepackage{subcaption}
\usepackage{amsthm}
\usepackage{multirow}
\usepackage{booktabs}
\usepackage{amsmath}
\usepackage[]{algorithm2e}
\usepackage{color}
\usepackage{multicol}
\usepackage{bm}
\usepackage{authblk}
\usepackage[capitalize]{cleveref}
\usepackage{titlesec}
\usepackage{cleveref}

\titlelabel{\thetitle.\;}

\crefname{section}{Sec.}{Secs.}
\Crefname{section}{Section}{Sections}
\Crefname{table}{Table}{Tables}
\crefname{table}{Tab.}{Tabs.}
\crefname{appendix}{Appx.}{Appxs.}
\crefname{figure}{Fig.}{Figs.}

\title{Learning to Break Deep Perceptual Hashing: The Use Case NeuralHash}

\makeatletter
\newcommand{\printfnsymbol}[1]{%
  \textsuperscript{\@fnsymbol{#1}}%
}
\makeatother

\author[1]{Lukas Struppek\thanks{Equal contributions. Contact: \textit{firstname.lastname@tu-darmstadt.de}}}
\author[1]{Dominik Hintersdorf\,\printfnsymbol{1}}
\author[2]{Daniel Neider}
\author[1,3]{Kristian Kersting}
\affil[1]{Department of Computer Science, TU Darmstadt, Darmstadt, Germany}
\affil[2]{Max Planck Institute for Software Systems, Kaiserslautern, Germany}
\affil[3]{Centre for Cognitive Science, TU Darmstadt, and Hessian Center for AI (hessian.AI)}

\begin{document}
\maketitle
\thispagestyle{empty}

%%%%%%%%% ABSTRACT
\begin{abstract}
Apple recently revealed its deep perceptual hashing system NeuralHash to detect child sexual abuse material (CSAM) on user devices before files are uploaded to its iCloud service.
Public criticism quickly arose regarding the protection of user privacy and the system's reliability. In this paper, we present the first comprehensive empirical analysis of deep perceptual hashing based on NeuralHash. Specifically, we show
that current deep perceptual hashing may not be robust. An adversary can manipulate the hash values by applying slight changes in images, either induced by gradient-based approaches or simply by performing standard image transformations, forcing or preventing hash collisions.
Such attacks permit malicious actors easily to exploit the detection system: from hiding abusive material to framing innocent users, everything is possible.
Moreover, using the hash values, inferences can still be made about the data stored on user devices. In our view, based on our results, deep perceptual hashing in its current form is generally not ready for robust client-side scanning and should not be used from a privacy perspective.\footnote{Published as a conference paper at the ACM Conference on Fairness, Accountability, and Transparency (FAccT) 2022. \href{10.1145/3531146.3533073}{https://doi.org/10.1145/3531146.3533073}.}
\end{abstract}

\section{Introduction}
In 2020, the US National Center for Missing \& Exploited Children (NCMEC) received over 21 million reports of online child sexual exploitation, an increase of 28\% from 2019~\cite{CyberTipline}. Child sexual abuse material (CSAM) is no niche phenomenon and must be pursued and prevented. With millions of users uploading pictures and other media to online platforms, there is an unmanageable flood of data. One approach to efficiently analyze this amount of data is to transform the images into a lower-dimensional space by extracting image-specific features. These compact representations enable an efficient search in and comparison of high-dimensional data. In this paper, we will focus on deep perceptual hashing that computes relatively short bit sequences of fixed-length, the so-called hashes or fingerprints, for multimedia content. Hashing generally describes a deterministic transformation of data into short bit sequences. Perceptual hashing aims to assign similar hashes to images with similar visual features. The more recent deep perceptual hashing algorithms utilize neural networks for feature extraction. Perceptual hashing algorithms differ significantly from traditional cryptographic hashing algorithms, which aim to produce widely differing hashes for minor input changes, the so-called avalanche effect.

Apple recently announced its NeuralHash~\cite{apple_tech_summary} system, a deep perceptual hashing algorithm for client-side content scanning. The approach focuses on identifying CSAM content in user files uploaded to Apple's iCloud service. Apple has made several assurances about privacy and security, such as a low risk of accounts being falsely flagged and restricting its access to private data. More detailed information on NeuralHash is provided in the official technical summary~\cite{apple_tech_summary}. According to Apple, only images uploaded to the iCloud servers will be hashed and compared against a hash database of known CSAM material. The hash databases are provided by NCMEC and other child protection agencies and are only available in encrypted form on the user devices. While Apple commissioned various expert opinions on the security of the system from independent researchers~\cite{Bellare, Forsyth, Pinkas}, there has been a lot of public criticism of the NeuralHash approach~\cite{snowden21, abelson2021bugs, lewis21_twitter}. Criticism is directed not only at possible privacy violations but also at the system's reliability. So far, however, there has been a lack of comprehensive analyses, at least publicly, of the core of the perceptual hashing process, the hash computation itself.

The European Union (EU) presented in 2020 their strategy~\cite{eu_csam_strategy} to become more effective in fighting child sexual abuse. The strategy specifically refers to end-to-end encryption and calls for technical solutions that allow companies to detect and report CSAM material transferred in encrypted communication systems. The subsequent EU regulation 2021/1232~\cite{eu_regulation}, which also faces criticism~\cite{chat_control_breyer}, establishes temporarily limited rules with the sole purpose of allowing service providers to use specific technologies for the processing of personal and other data to the extent necessary to detect and report CSAM material. Without the regulation stating specific technical details, it can be assumed that hash-based client-side scanning approaches could be a way to achieve this. Similarly, the UK governments' Online Safety Bill~\cite{draft_online_safety_bill} suggests a comparable direction and establishes new ways to regulate online content to prevent the distribution of harmful material.

A recent paper~\cite{abelson2021bugs} by well-known cybersecurity researchers and encryption system inventors, such as Hal Abelson, Carmela Troncoso, and Josh Benaloh, outlines and critiques the potential security and privacy risks of client-side scanning technologies. The authors argue that even if the initial goal of these systems is only the detection of clearly illegal content, tremendous pressure to expand the scope of application will arise with time. People would then have little chance to resist the expansion of the system or prevent its abuse.

In this paper, we investigate the perceptual hashing components of NeuralHash, in particular, the embedding neural network and the hashing step. It might be to some extent common knowledge that neural networks are susceptible to various kinds of attacks. However, we are convinced that it is important to demonstrate that this susceptibility is not only interesting from a researcher's point of view but actually affects systems used by millions of users who might not be aware of the risks of these systems. We further want to emphasize that by using neural networks and, therefore, being able to calculate the gradients with respect to the inputs, most of the attacks are rather easy to perform, exposing various risks to manipulate the systems.

Our focus lies on NeuralHash because it is the first prominent representative to shift content detection from the server-side to user devices. This approach poses additional major risks and insecurities, such as causeless algorithmic surveillance of users. We show that deep perceptual hashing has various downsides when applied to real-world large-scale image detection.
Our research aims to point out drawbacks of deep perceptual image hashing, support the development of more robust systems and encourage a discussion on the general deployment of this technology. NeuralHash merely acts as a current real-world example in this case.

We proceed as follows. In Section~\ref{sec:background}, we first generally introduce deep perceptual hashing together with the specifics of NeuralHash and a general overview of attacks against neural networks. Our first adversarial setting in Section~\ref{sec:adversary_1} then sheds light on the possibility to create hash collisions with minor visual image changes. In Sections~\ref{sec:adversary_2} and~\ref{sec:adversary_3}, we show that the hash computation is susceptible to gradient-based and gradient-free image transformations. In our fourth adversarial setting in Section~\ref{sec:adversary_4}, we demonstrate that a hash value itself actually contains information about its corresponding image and might leak information about the content on user devices. We conclude our work by a discussion of our results and their implications in Section~\ref{sec:lessons}, followed by a summary in Section~\ref{sec:conclusion}.

Before diving into the details, we want to make the following two statements:
\begin{enumerate}
    \item We explicitly condemn the creation, possession, and distribution of child pornography and abusive material and strongly support the prosecution of related crimes. With this work, we in no way intend to provide instructions on how to bypass or manipulate CSAM filters. In turn, we want to initiate a well-founded discussion about the effectiveness and the general application of client-side scanning based on deep perceptual hashing.
    \item We have no intention to harm Apple Inc. itself or their intention to stop the distribution of CSAM material. NeuralHash merely forms the empirical basis of our work to critically examine perceptual hashing methods and the risks they may induce in real-world scenarios.
\end{enumerate}

\begin{figure*}[ht!]
\centering
\includegraphics[width=\linewidth]{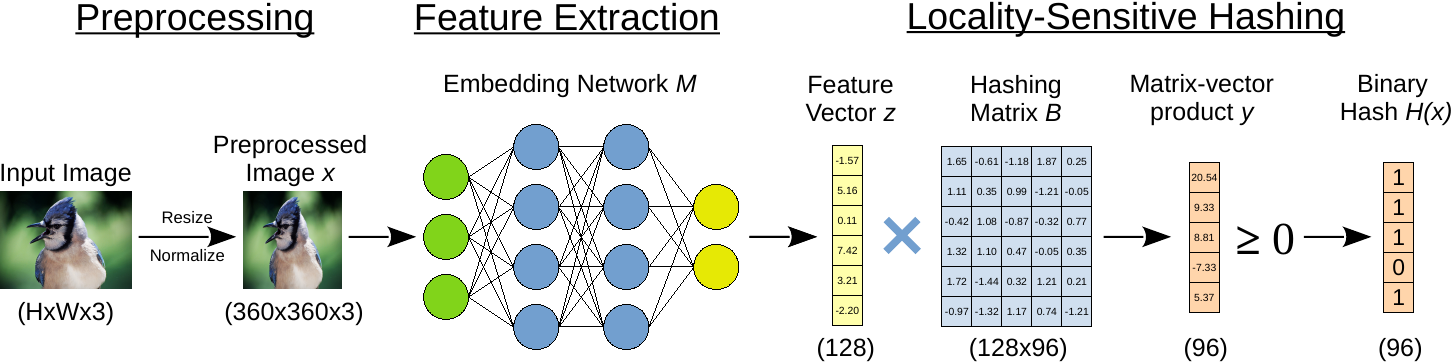}
\caption{NeuralHash pipeline as deployed on user devices. The pipeline consists of an embedding network and a locality-sensitive hashing (LSH) step. The embedding network maps the preprocessed images into an abstract feature representation vector. LSH then maps each vector into a specific bucket by checking its position relative to the hyperplanes defined in the hashing matrix.
\label{fig:neural_hash_architecture}}
\end{figure*}

\section{Background}\label{sec:background}
We start by formally defining a perceptual hashing system, describing the characteristics of NeuralHash, and giving an overview of security-related attacks against neural networks.

\subsection{Perceptual Hashing and NeuralHash}
Perceptual hashing algorithms, e.g., Apple's NeuralHash~\cite{apple_tech_summary}, Microsoft's PhotoDNA~\cite{photodna} and Facebook's PDQ~\cite{pdq}, aim to compute similar hashes for images with similar contents and more divergent hashes for different contents. In recent years, various deep hashing algorithms based on convolutional neural networks have been proposed~\cite{Liong, Haomiao, Dayan, Zhao2015DeepSR}. They all rely on deep neural networks to first extract unique features from an image and then compute a hash value based on these features. In the following, we introduce the basic deep perceptual hashing approach, before describing the specifics of NeuralHash. Figure~\ref{fig:neural_hash_architecture} gives an overview of the hash computation steps, which we now explain in detail.

We define a (perceptual) hash function $H \colon \mathbb{R}^{H\times W\times C} \to \{0,1\}^k$ that maps an image $x$ to a $k$-bit binary hash. Let $h_i(x)$ further denote the partial hash function for the $i$-th hash bit, i.e., $H(x) = \bigl( h_1(x), \ldots, h_k(x) \bigr)$. Perceptual hashing algorithms usually consist of two components. First, a shared feature extractor $M(x)\colon \mathbb{R}^{H \times W \times C} \to \mathbb{R}^{m}$ extracts visual features from an image $x$ and encodes them in a feature vector $z \in \mathbb{R}^{m}$. This resulting feature vector $z$ is an abstract numeric interpretation of the image's characteristic features.

\begin{figure}[h]
\centering
\includegraphics[width=0.78\linewidth]{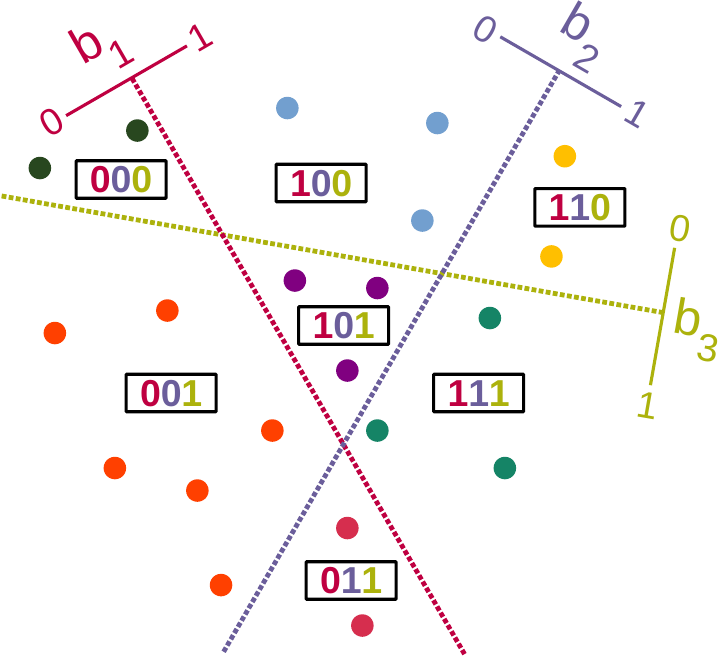}
\caption{ Locality-sensitive hashing (LSH) scheme. Each hyperplane divides the space into two parts, with a bit state \{0, 1\} assigned to each side. The polytopes constrained by the hyperplanes are called buckets. Each bucket is assigned a unique binary hash code based on its relative position to each hyperplane. All data points in the same bucket are assigned the same hash code.
\label{fig:lsh}}
\end{figure}

Next, locality-sensitive hashing (LSH)~\cite{IndykM98, Gionis99} is used to assign close feature vectors to buckets with similar hash values. Among other LSH methods, random projection can be used to quickly convert the extracted features into a bit representation. For each of the $k$ bits, a (random) hyperplane is defined in the hashing matrix $B \in \mathbb{R}^{m \times k}$.
The bit value $h_i(x)$ is set by checking on which side of the $i$-th hyperplane feature vector $z$ lies. In practice, the feature vector $z$ is first transformed into $y=B \cdot z$ with ${y \in \mathbb{R}^{k}}$ by calculating the product of the hashing matrix and the embedding. The real-valued vector $y$ is then finally converted to a bit vector by applying a Heaviside step function to each element $y_i$. The result is a binary hash vector $H(x)$ containing $k$ bits. For the figures in this paper, we represent the computed hashes in their equivalent, compact hexadecimal representation. We visualized the basic concept of LSH in a 2D case with three random hyperplanes in Figure~\ref{fig:lsh}. Points with the same color are assigned to the same bucket and result in the same hash.

To compare similarity and distance between two hashes, we define the normalized Hamming distance of two binary hashes $x$ and $y$ as $\delta(H(x), H(y))=\frac{1}{k} \sum\nolimits_{i=1}^k|h_i(x)-h_i(y)|$. In other words, $\delta$ describes what percentage of bits in the two hashes differ. It allows quantifying the distance between two hashes and analyzing their proximity. For brevity, we refer to it simply as the Hamming distance.

Figure~\ref{fig:neural_hash_architecture} illustrates the algorithmic pipeline of NeuralHash, which we now describe in more detail. NeuralHash~\cite{apple_tech_summary} expects resized RGB input images with shape $360 \times 360\times 3$ within the pixel range $[-1, 1]$.
The embedding network $M$ is based on a modified MobileNetV3~\cite{mobilenetv3} architecture with 1.8 million parameters. The network has been trained in a self-supervised training scheme with a contrastive loss. The idea behind contrastive learning~\cite{hadsell} is to create a positive and negative pair of samples for each training image, the so-called anchors. For the positive pair, the anchor is a transformed version of the training image so that it remains perceptually similar, e.g., by scaling or cropping. The negative pair is built up by selecting a perceptually different image, e.g., from another class. A neural network is then trained to output close feature vectors $z$ for positive pairs and more distant vectors for negative pairs.

Closeness is usually described by the cosine of the angles between the feature vectors. Unfortunately, Apple does not state any additional details on training data, loss functions, or hyperparameters. The embedding network produces feature vectors $z$ of length $m=128$. The hashing matrix $B$ has shape $128 \times 96$ and, consequently, defines 96 hyperplanes. Therefore, the final hashes have a fixed length of $k=96$ bits. For our experiments, we extracted the neural network as well as the hashing matrix used by Apple from a Mac computer. More details about the extraction can be found in Appendix~\ref{app:experimental_details}.

\subsection{Attacks Against Neural Networks}
To put our work into context, we briefly introduce common attacks against neural networks. Most similar to our gradient-based hash attacks in Sections~\ref{sec:adversary_1} and \ref{sec:adversary_2}, are adversarial attacks~\cite{adv_attacks}. They apply precisely tuned perturbations to images that are hardly perceivable to humans but lead to misclassifications. Most adversarial attacks compute the gradients of the output with respect to a model's input and perform one~\cite{fgsm} or many~\cite{pgd} optimization steps to force the model's outputs to specific values.

While much work has been done on the general vulnerability of neural networks to adversarial attacks, little work has been published on the vulnerability of deep hashing functions. As a first step, targeted~\cite{bai2020targeted, Wang21, Xiao_2021_CVPR, Wang_2021_CVPR} and untargeted~\cite{Yang20} adversarial attacks against deep hashing-based retrieval systems were proposed. The goal of these attacks is to manipulate input images in a way that the target system retrieves objects with a specific target label or objects that are semantically irrelevant to the original inputs.

Another work~\cite{DolhanskyCollisions} demonstrated the susceptibility of various image hashing functions against gradient-based collision attacks. Similar works investigated the robustness of non-deep perceptual hashing algorithms against adversarial attacks~\cite{hao21, jain21} and their robustness~\cite{eval_robustness_drmic} against visible image modifications. Other research has focused on recovering the original images from the supposedly anonymous real-valued image hashes by training a deep neural network to reconstruct inputs, given their hash values~\cite{RevHashNet_wang}. However, they only investigated the de-hashing of non-deep perceptual hashing algorithms.

Until now there is no comprehensive work on the technical vulnerabilities from a machine learning perspective of NeuralHash or client-side scanning based on deep perceptual hashing yet. However, a few proof-of-concept implementations to create hash collisions on NeuralHash exist~\cite{neuralhashcollider, neuralhashcollider2, kilcher}. Our work aims to fill this research gap and help to weigh the privacy benefits and potential threats.

\section{Adversary 1 -- Hash Collision Attacks}\label{sec:adversary_1}
\begin{figure*}[t]
\centering
\includegraphics[width=\linewidth]{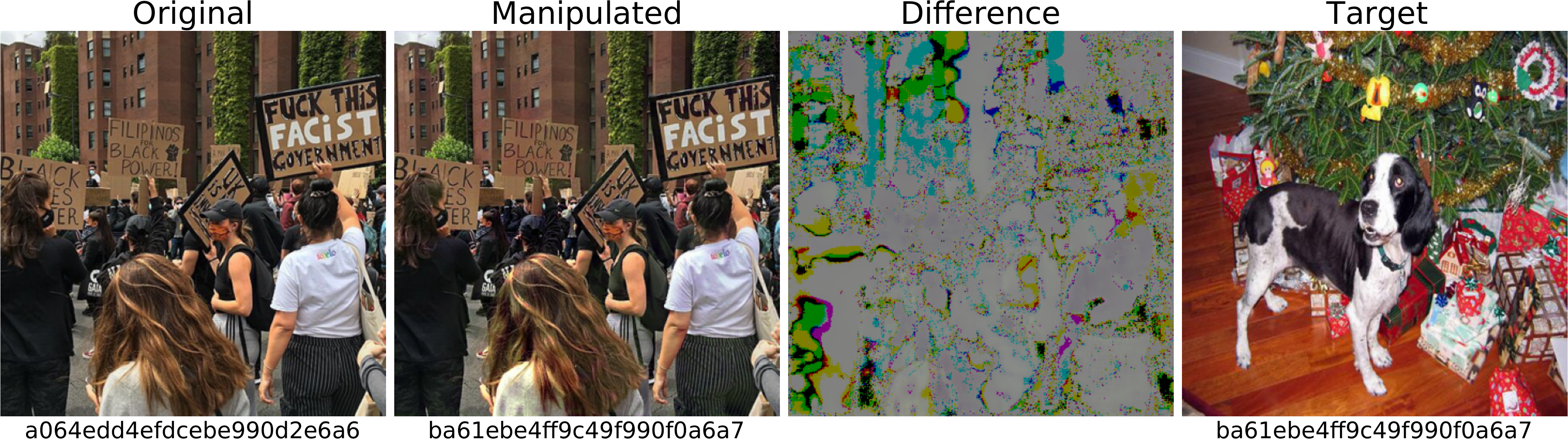}
\caption{We manipulated the original image~\cite{wikimediaProtest} to have the same hash as the target image (Adversary 1). The manipulated image is visually hardly distinguishable from the original since the (normalized) differences are small. Still, the manipulated image is assigned the same hash as the visually completely different target image. This demonstrates the practicability and danger of hash collision attacks.}
\label{fig:hash_poisoning}
\end{figure*}
In our first adversarial setting, we investigate the creation of hash collisions. We perturb images so that their computed hashes match predefined target hashes. This issue holds several explosive scenarios in reality. Given a set of target hashes from the (CSAM) hash database, an adversary can create fake images whose hashes match those in the database without containing any sensitive material at all. Distributed across many devices, this may lead to a large number of false-positive alarms in the (possibly partly human) detection system and, as a consequence, result in the framing of innocent users or distributed denial of service (DDoS) attacks.

Another worrisome scenario could be that service providers, governments, or other powerful organizations either add additional hashes to the database or manipulate ordinary images that show political or social content that is undesirable to them, such as government critics or LGBTQ+ supporters. By spreading such manipulated images through social media, they might end up on the devices of supporters of these groups. Identification, surveillance, and persecution of these people are then possible by detecting hash collisions with the database. This poses a great danger, especially for people in states with restricted fundamental rights or totalitarian regimes.

Even without access to the hash database, which governments, after all, might have, an adversary could simply collect its own CSAM material and compute its hashes. Assuming a large enough database, the adversary could obtain, at least, some share of the true database hashes. Figure~\ref{fig:hash_poisoning} shows a manipulated image from a protest march that results in the same hash value as a non-related target image. For this paper, we created a surrogate hash database with dog images that acts as a list of images that should be detected by the system.

\textbf{Technical Realization.} To create hash collisions, we first computed the hash of an input image $x_\mathit{orig}$ and took the target hash $\hat{H}$ from the (surrogate) hash database with the smallest Hamming distance to the computed hash of $x_\mathit{orig}$. After defining the target hash $\hat{H}=(\hat{h}_1, \ldots, \hat{h}_k)$, we directly optimized the input image $x=x_\mathit{orig}$ such that its binary hash value approaches $\hat{H}$. For this task, we defined a Hinge loss
\begin{equation}\label{eq:hinge}
\begin{aligned}
    \mathcal{L}_\mathit{Hinge}(x, \hat{H})= \frac{1}{k} \sum\nolimits_{i=1}^k max\{0, d-y_i \cdot \psi(\hat{h}_i)\} \\
\end{aligned}
\end{equation}
with $d \geq 0$ describing the margin to the hyperplanes and $y = B \cdot M(x)$ with $y\in \mathbb{R}^{96}$ being the real-valued hash output before binarization. The operation $\psi(\hat{h}_i)=\mathit{sign}(\hat{h}_i-0.5)$ replaces each $0$-bit in the hash vector by $-1$. In our experiments, we set $d=0$ to optimize only until the hash value of the optimized image matches $\hat{H}$. This is achieved when the signs of $y_i$ and $\psi(\hat{h}_i)$ match at all $k$ positions. By setting $d>0$, the distance to the LSH hyperplanes could be increased, leading to more robust image hashes.

Furthermore, we added a structural similarity (SSIM)~\cite{ssim} penalty term to reduce visual conspicuities between an optimized image and its original counterpart.
The structural similarity (SSIM) is defined as
\begin{equation}\label{eq:ssim}
    \mathit{SSIM}(x, y) = \frac{(2\mu_x\mu_y+C_1)(2\sigma_{xy}+C_2)}{(\mu_x^2+\mu_y^2+C_1)(\sigma_x^2+\sigma_y^2+C_2)}.
\end{equation}

\begin{figure}
\centering
\includegraphics[width=0.99\linewidth]{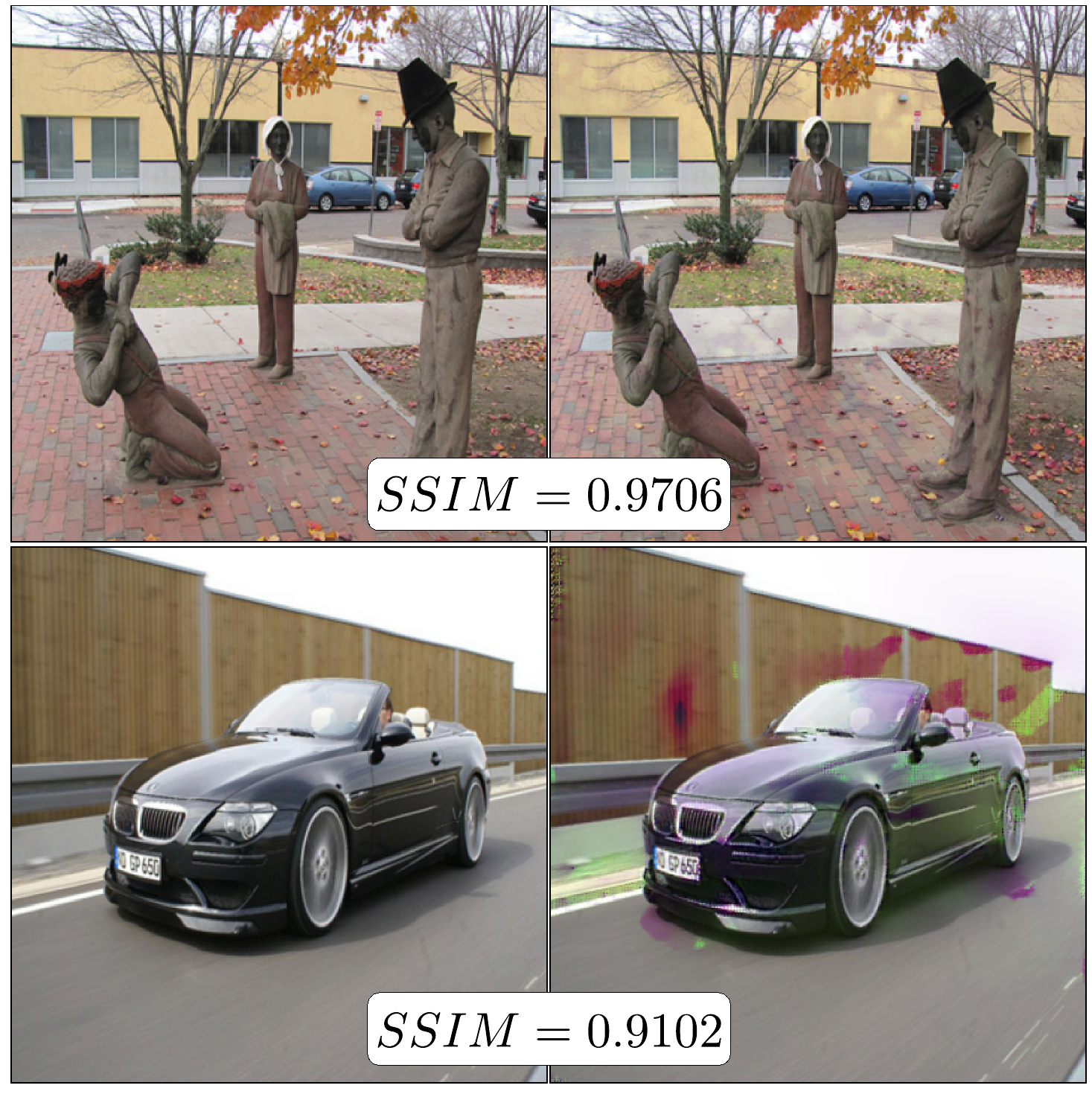}
\caption{Additional collision results with their SSIM values between the original images (left) and the manipulated ones (right).
\label{fig:ssim_examples}}
\end{figure}

For two images $x$ and $y$, the parameters $\mu_i$ and $\sigma_i^2$ denote the mean and variance of each image's pixels. Further, $\sigma_{xy}$ denotes the covariance of $x$ and $y$. The constants $C_1$ and $C_2$ are added for numerical stability and are set to $C_1=10^{-4}$ and $C_2=9\cdot 10^{-4}$, respectively, in our experiments. The closer $\mathit{SSIM}\in [0, 1]$ is to 1, the more similar $x$ is to the original image without perturbations. We computed the SSIM, weighted by parameter $\lambda$, in each optimization step between the optimized image $x$ and its unmodified counterpart $x_\mathit{orig}$ to improve the image quality. To get an intuition for SSIM, we state two collision results together with their SSIM values in Fig.~\ref{fig:ssim_examples}.

In total, we attempted to solve the following optimization problem:
\begin{equation} \label{eq:adversary1_problem}
\begin{split}
\min_x & \quad \mathcal{L}_\mathit{Hinge}(x, \hat{H}) - \lambda \cdot \mathit{SSIM}(x, x_\mathit{orig}) \\
\textrm{s.t.} & \quad H(x) = \hat{H} \\
 & \quad x\in[-1, 1]^{H \times W \times C}.
\end{split}
\end{equation}

\textbf{Experimental Setup}; see Appendix~\ref{app:adv1_details} for additional details. For our analyses, we created a surrogate hash database by hashing all 20,580 dog images from the Stanford Dogs dataset~\cite{dogs_dataset}. We then performed our attacks by modifying the samples on the first 10,000 samples from the ImageNet ILSVRC2012~\cite{deng2009imagenet, ILSVRC15} test split.

We used the Adam optimizer~\cite{adam} to directly optimize $x$. We further set $\lambda=100$ and stopped the optimization when either $H(x)=\hat{H}$ was satisfied or aborted after 10,000 iterations. If we did not stop the optimization process when $H(x)=\hat{H}$ is fulfilled, the image quality might be further improved due to the SSIM term.

\textbf{Results.} Table~\ref{tab:hash_collision_results} states our collision attack results. The success rate (SR) indicates the share of images whose hashes have successfully been changed. We further state the mean $\ell_2$ and $\ell_\infty$ distances between $x_\mathit{orig}$ and its optimized counterpart $x$ to quantify pixel-wise image changes. We also computed the mean SSIM values to take the image quality into account. Steps denote the mean number of optimization steps performed until a hash collision occurred.

\begin{table}[t]
    \centering
    \resizebox{\columnwidth}{!}{
    \begin{tabular}{ccccc}
\textbf{SR} & $\bm{\ell_2}$ & $\bm{\ell_\infty}$ & \textbf{SSIM}  & \textbf{Steps} \\
    \toprule
    $90.81\%$ & $20.8136\pm7.97$ & $0.3120\pm0.22$ & $0.9647\pm0.03$ & $1190\pm1435$ \\
    \bottomrule
    \end{tabular}
    }
  \caption{Evaluation metrics for our hash collision attack computed on an ImageNet subset. For all values, we state the mean and the standard deviation.}
  \label{tab:hash_collision_results}
\end{table}

We could force hash collisions in about 90\% of all images, demonstrating the real applicability of the attack. The visual salience of the modifications varies in strength. Most manipulated images contain some patches of color, which are sometimes conspicuous but often hardly noticeable. Figure~\ref{fig:hash_poisoning} shows such an example where the visual differences between the original and manipulated images are small. We state further collision examples in Figure~\ref{fig:ssim_examples} and Appendix~\ref{app:add_hash_collision_samples}. We argue that by tuning the attack's hyperparameters and a careful selection of input images, success rate and visual indistinguishability between the images might be further improved in real scenarios. We also investigated how pretrained GANs can be used to generate images with specific hashes in Appendix~\ref{app:adv1_gan}.

In summary, we demonstrated that hash collisions can easily be forced in NeuralHash and might build the base for serious attacks targeting the service provider or, even worse, persecution of political opponents. While some of the introduced image changes are visible, they are barely noticeable in most images, as also the SSIM values indicate.

\section{Adversary 2 -- Gradient-Based Evasion
\mbox{Attacks}}
\label{sec:adversary_2}
Next, we investigate how robust NeuralHash is and if we can change the hash of any image by perturbating it. This is also called a detection evasion attack and aims to evade detection of sensitive material through hash comparison by an automated system. A perceptual hashing algorithm used to identify sensitive material should be robust to small image changes to provide reliable results.

Two kinds of image manipulations are plausible: optimization-based approaches that compute image-specific perturbations and transformation-based approaches that transform an image using simple procedures provided by standard image editors. Non-robust perceptual hashing algorithms would make it easy to hide sensitive material from detection and call into question the overall effectiveness of such systems.

In this section, we follow an optimization-based approach and compute gradient-based perturbations that aim to alter the hash of an image with minimal differences from the original one. We propose three different attacks: Standard, Edges-Only, and Few-Pixels. Our Standard attack allows changes to all parts of an image. The Edges-Only attack first detects the edges in an image and then restricts the manipulation to the edge pixels. For the Few-Pixels attack, we attempt to change as few pixels of an image as possible.

\begin{figure*}[h]
\centering
\includegraphics[width=0.78\linewidth]{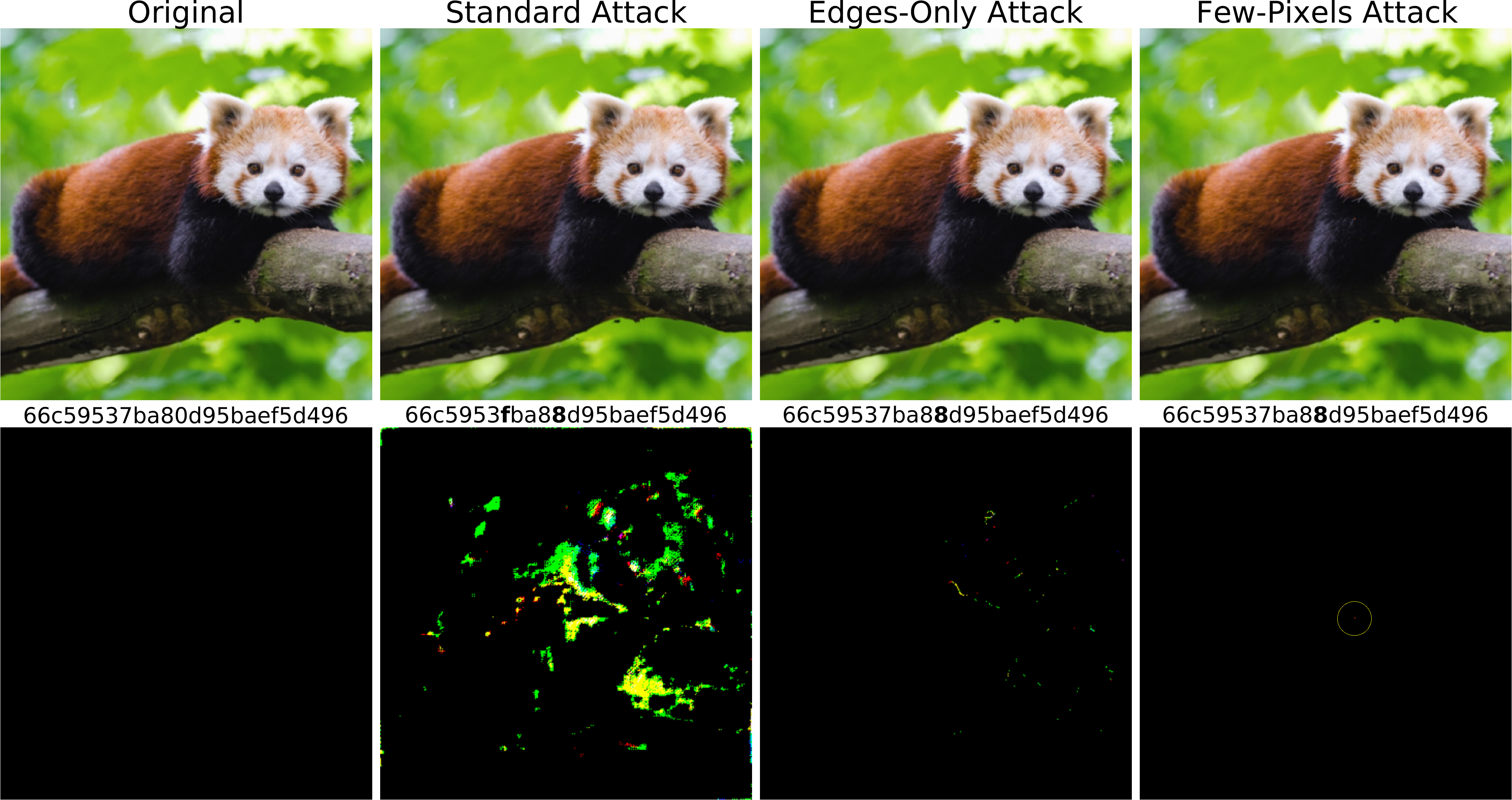}
\caption{Visualization of our gradient-based evasion attacks (Adversary 2). The added perturbations are hardly visible, even changing a single pixel leads to a hash change. Normalized differences are visualized below. Here, black marks image parts that did not have been modified. Zoom in for visual details.}
\label{fig:attack_example}
\end{figure*}

\textbf{Technical Realization.} To change the original hash $\tilde{H}=H(x_{orig})=(\tilde{h}_1, \ldots \tilde{h}_k)$ of a given input image $x_\mathit{orig}$, we used a negative mean squared error (MSE) loss to increase the hash discrepancy between an image $x_\mathit{orig}$ and its manipulated counterpart $x$ as
\begin{equation}
    \mathcal{L}_\mathit{MSE}(x, \tilde{H}) = - \frac{1}{k} \sum\nolimits_{i=1}^k \left( \sigma(c\, \widetilde{y_i}) - \tilde{h}_i \right)^2 .
\end{equation}
Instead of $\mathcal{L}_\mathit{MSE}$, $-\mathcal{L}_\mathit{MSE}$ is reducing the Hamming distance between two hashes and can therefore be used to force hash collisions~\cite{neuralhashcollider}.

We make the binarization step differentiable by replacing it with a sigmoid function $\sigma(x)=\frac{1}{1+e^{-x}}$. To push the sigmoid values closer to zero and one, we scaled its inputs by factor $c$. In our experiments, we set $c=5$. For a stable optimization, we normalized the matrix-vector product $y=B\cdot M(x)$ by $\widetilde{y}=\frac{y}{max(\|y\|_2, \epsilon)}$ and set $\epsilon=10^{-12}$ for numerical stability. For each attack, we optimized until a minimal Hamming distance $\delta_0$ between $x$ and $x_\mathit{orig}$ is exceeded. As with Adversary 1, we also added an SSIM penalty term weighted by hyperparameter $\lambda$ to reduce visual conspicuities.

The full optimization goal can be formulated as
\begin{equation}
\begin{split}
    \min_x  & \quad \mathcal{L}_\mathit{MSE}(x, \tilde{H}) - \lambda * \mathit{SSIM}(x, x_\mathit{orig}) \\
    \textrm{s.t.} & \quad \delta(H(x),\tilde{H})>\delta_0 \\
    & \quad x\in[-1, 1]^{H \times W \times C}.
\end{split}
\end{equation}

For the Edges-Only attack, we first applied a Canny edge detector~\cite{canny_edges} in the greyscale version of an image. We then only allowed changes in the set of edge pixels by applying a binary mask to the gradients during optimization.

As for our Few-Pixels attack, we started by selecting a single pixel to optimize by taking the pixel with the highest absolute $\mathcal{L}_\mathit{MSE}$ loss gradient value across all three color channels. We then again applied a binary mask to the gradients and optimized for $N$ steps. If the hash did not change, we added another pixel to the pixel set and again optimized for $N$ steps on all pixels of the set. Additional pixels were again selected by their absolute $\mathcal{L}_\mathit{MSE}$ loss gradient values. We repeatedly added more pixels and modified the set of pixels until there was a change in the hash value or a threshold of the number of pixels has been reached.

\textbf{Experimental Setup}; see Appendix~\ref{app:adv2_details} for additional details. We performed our evasion attacks on the first 10,000 ImageNet test samples. We used the Adam optimizer to directly optimize an image $x$. We applied a decaying weight for the SSIM term and set $\lambda=5\cdot 0.99^{step}$, where $step$ denotes the number of optimization steps already performed. By decaying the weight, we avoided that the optimizer getting stuck in local minima in some cases. For the Standard and Edges-Only attacks, we performed a maximum of 1,000 optimization steps.

In the Few-Pixels attack, we set $\lambda=0$ and, therefore, removed the SSIM term. The optimization is aborted after 150 pixels without a hash change. In all three attacks, we set the minimum Hamming distance $\delta_0=0$ and, consequently, stopped each attack when $H(x) \neq \tilde{H}$.

We also can force our attacks to produce larger perturbations and move a sample even further away from its original hash in terms of the Hamming distance by setting $\delta_0>0$. We repeated our Standard attack with minimal Hamming distances $\delta_0 \in [0, 0.5]$ and increased $\delta_0$ in steps of $0.05$.

\begin{figure*}[ht]
\centering
\includegraphics[width=0.95\linewidth]{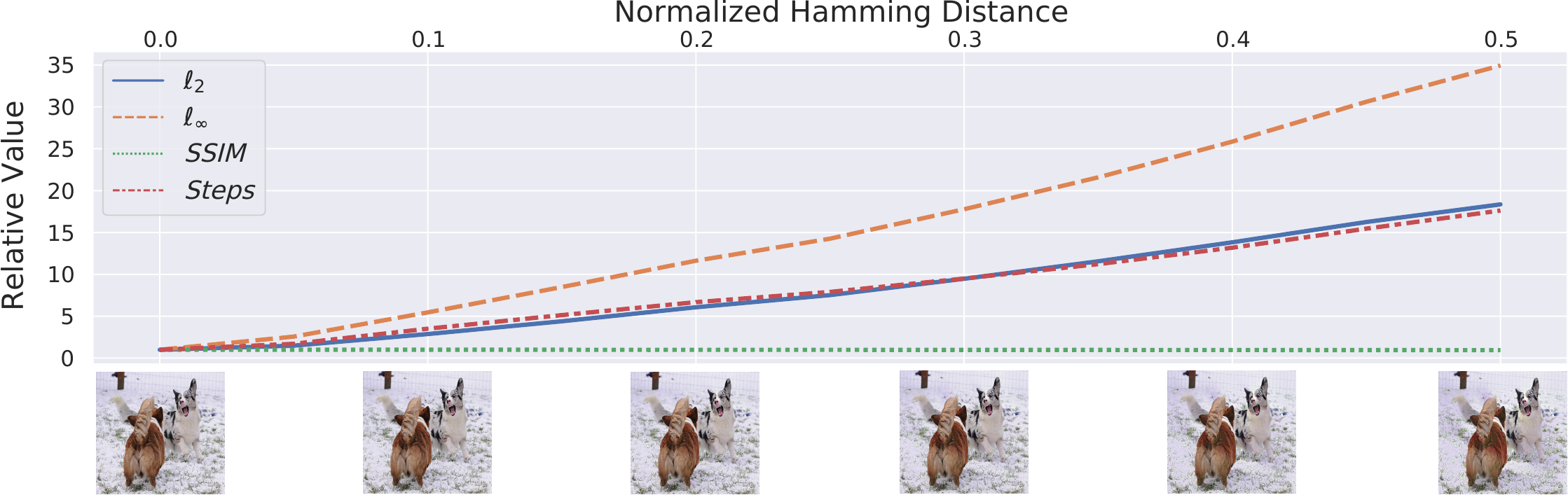}
\caption{We performed the Standard evasion attack (Adversary 2) with a varying minimum Hamming distance $\delta_0$, which can also be interpreted as the minimum percentage of hash bits to flip. We report the metrics relative to the results for $\delta_0=0$, as stated in Table~\ref{tab:change_hash_results}. For example, an $\ell_2$ value of $10$ means a ten times higher average $\ell_2$ distance between $x_\mathit{orig}$ and the manipulated image compared to the average distance when the first bit flips. Below the graph, we state the optimization results for a random image at increasing Hamming distances.}
\label{fig:varying_hamming_distances}
\end{figure*}

\textbf{Results.} Table~\ref{tab:change_hash_results} states the results for our gradient-based hash change attacks with $\delta_0=0$. We rely on the same metrics as used for our Adversary 1. The success rate (SR) indicates the share of images whose hashes could be changed before the maximum number of iterations or pixels has been reached. All three attacks were able to change the hash value of an image in the large majority of the cases. While the Standard Attack changed the hashes of all images, the Edges-Only and Few-Pixels attacks only failed in a few cases. Probably the optimization space in these attacks was too restricted, and the restrictions could be relaxed to achieve higher success rates.

The distances between the original and manipulated images are quite low, and the mean SSIM values are close to one, demonstrating that minor perturbations are sufficient to alter an image's hash. Figure~\ref{fig:attack_example} illustrates the effectiveness of our attacks and shows that the changes are hardly visually perceivable. We state additional samples for each attack in Appendix~\ref{app:add_hash_change_samples}.

\begin{table}[t]
    \centering
     \resizebox{\columnwidth}{!}{
    \begin{tabular}{lcccc}
    \textbf{Attack}             & \textbf{Standard}     & \textbf{Edges-Only}       & \textbf{Few-Pixels} \\
    \toprule
    \textbf{SR}                 & $100.00\%$            & $99.95\%$                 & $98.21\%$     \\
    $\bm{\ell_2}$               & $0.7188\pm0.28$       & $1.3882\pm1.37$           & $2.9100\pm2.06$ \\
    $\bm{\ell_{\infty}}$        & $0.0044\pm0.00$       & $0.0841\pm0.07$           & $0.8298\pm0.25$ \\
    \textbf{SSIM}               & $0.9999\pm0.00$       & $0.9996\pm0.00$           & $0.9989\pm0.00$  \\
    \textbf{Steps}              & $\phantom{00}5.40\pm4.98$       & $\phantom{00}150.24\pm113.96$       & $\phantom{.}323.01\pm3901$ \\
    \bottomrule
    \end{tabular}}
  \caption{Evaluation metrics (mean + standard deviation) for our three gradient-based evasion attacks computed on an ImageNet subset.}
  \label{tab:change_hash_results}
\end{table}

The Standard attack only needs about five optimization steps on average to force at least a single hash bit to flip. In the Few-Pixels attack, only 21.5 pixels were changed on average, $0.017\%$ of all pixels. All gradient-based attacks in this and the previous section can also be applied to images before resizing them into a square since most resizing operations, such as bilinear interpolation, are differentiable. We illustrate this fact with a few samples without resizing in Appendix~\ref{app:add_hash_change_samples}.

We plot the Standard attack results for increasing $\delta_0$ in Figure~\ref{fig:varying_hamming_distances}. While the number of optimization steps required and the $\ell_2$ and $\ell_\infty$ distances increase proportionally, the SSIM only decrease slightly, 0.96 on average for $\delta_0=0.5$, a flip of more than half of the bits, after all, indicating that even for larger Hamming distances the visual quality decreases only slightly. For all values of $\delta_0$, we achieved a success rate of 100\%.

Our results in this adversarial setting demonstrate that hash changes can easily be forced by gradient-based attacks. For humans, the induced changes are hardly perceivable. It is also possible to limit image manipulations to edges and even single pixels to keep the number of changes small. Our analysis shows that NeuralHash is in no way robust against gradient-based evasion attacks and questions the general robustness of neural network-based hashing approaches.

\begin{figure*}[h]
    \centering
    \begin{subfigure}[t]{0.325\textwidth}
        \includegraphics[width=\linewidth]{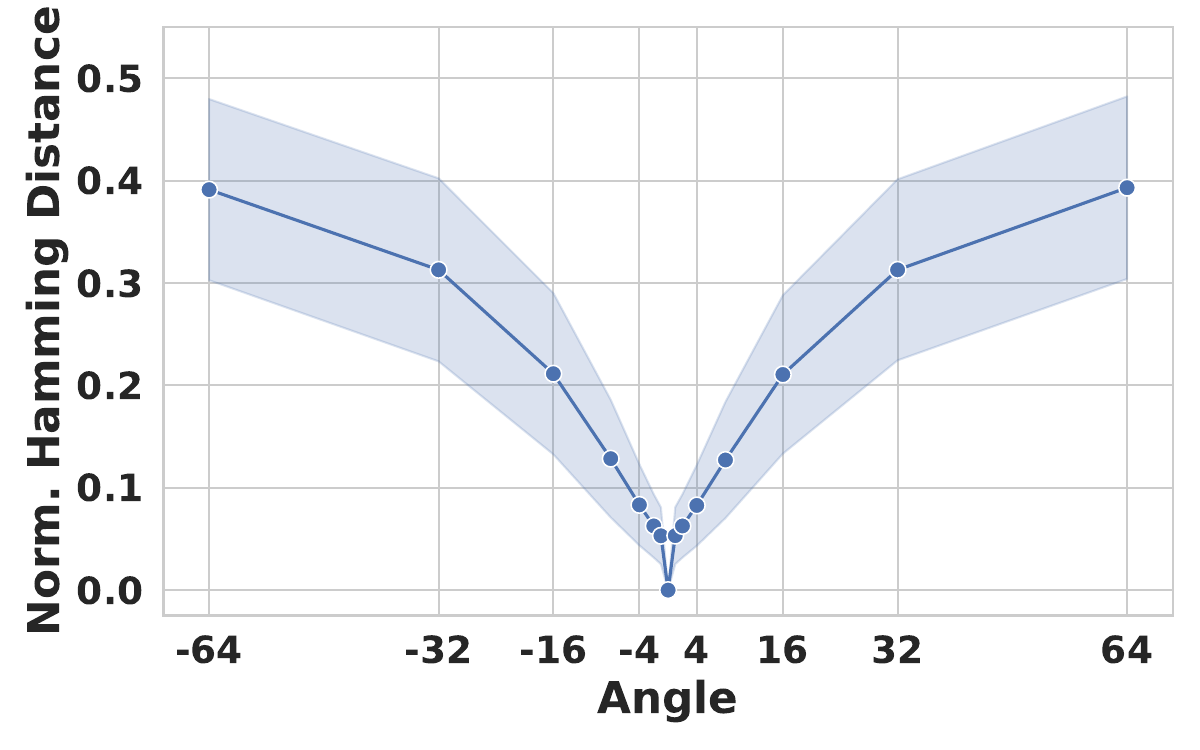}
        \caption{Image Rotation}
        \label{fig:rotation_robustness}
    \end{subfigure}
    \begin{subfigure}[t]{0.325\textwidth}
        \includegraphics[width=\linewidth]{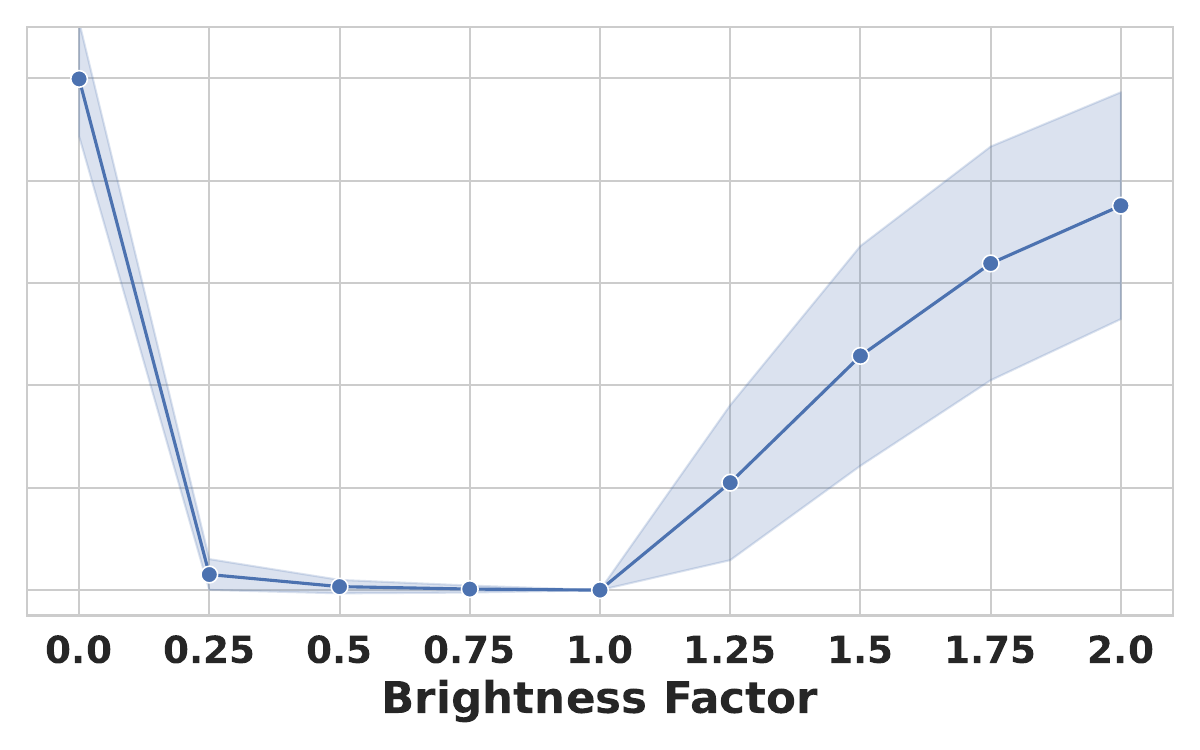}
        \caption{Brightness Changes}
        \label{fig:brightness_robustness}
    \end{subfigure}
    \begin{subfigure}[t]{0.325\textwidth}
        \includegraphics[width=\linewidth]{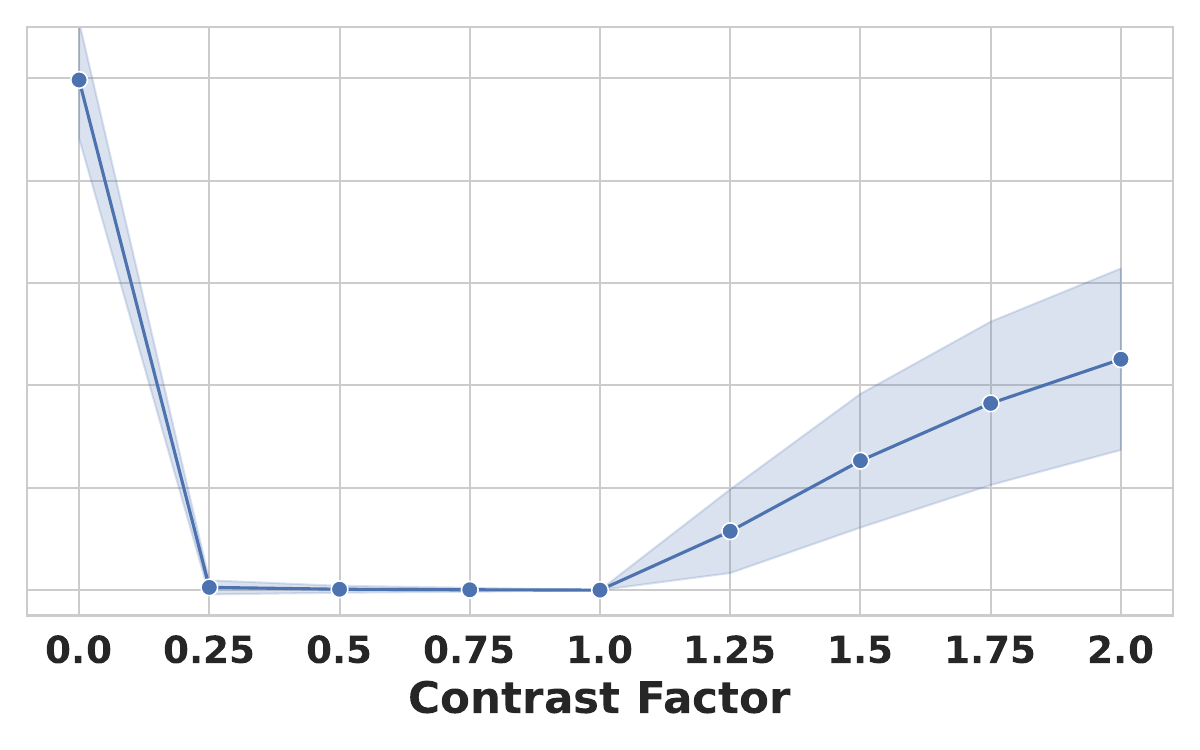}
        \caption{Contrast Changes}
        \label{fig:contrast_robustness}
    \end{subfigure}
    \begin{subfigure}[b]{0.325\textwidth}
        \includegraphics[width=\linewidth]{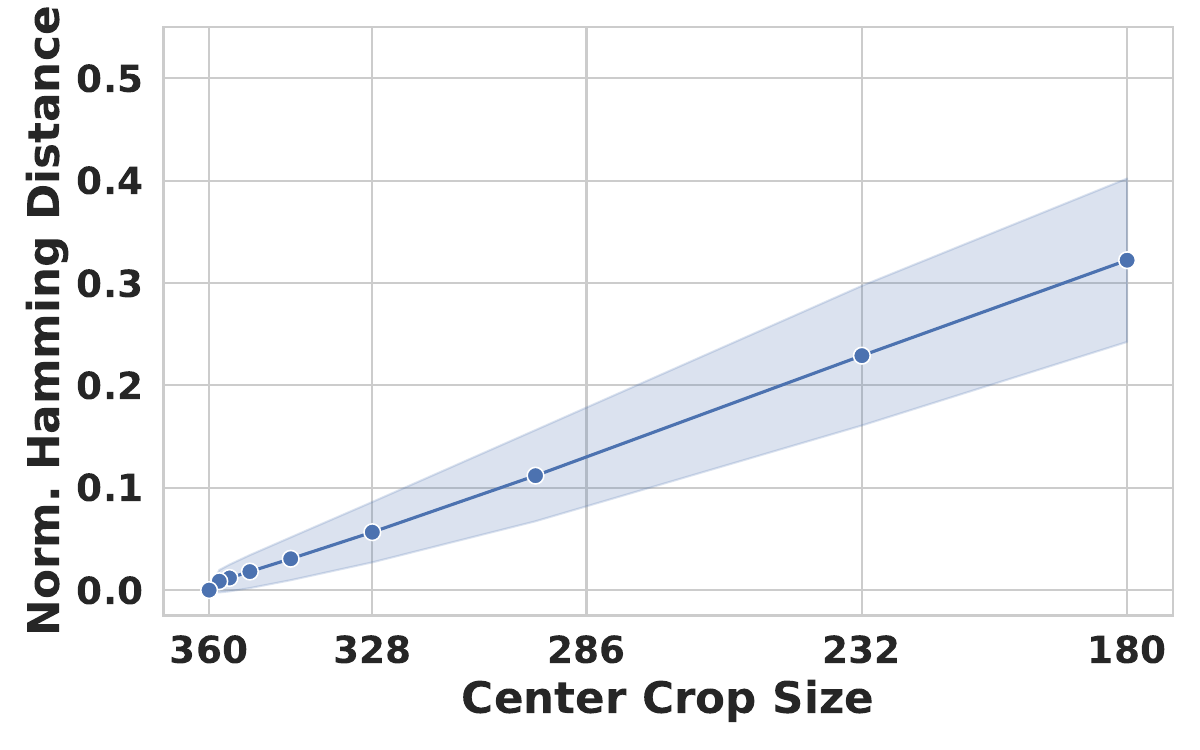}
        \caption{Center Cropping}
        \label{fig:crop_robustness}
    \end{subfigure}
    \begin{subfigure}[b]{0.325\textwidth}
        \includegraphics[width=\linewidth]{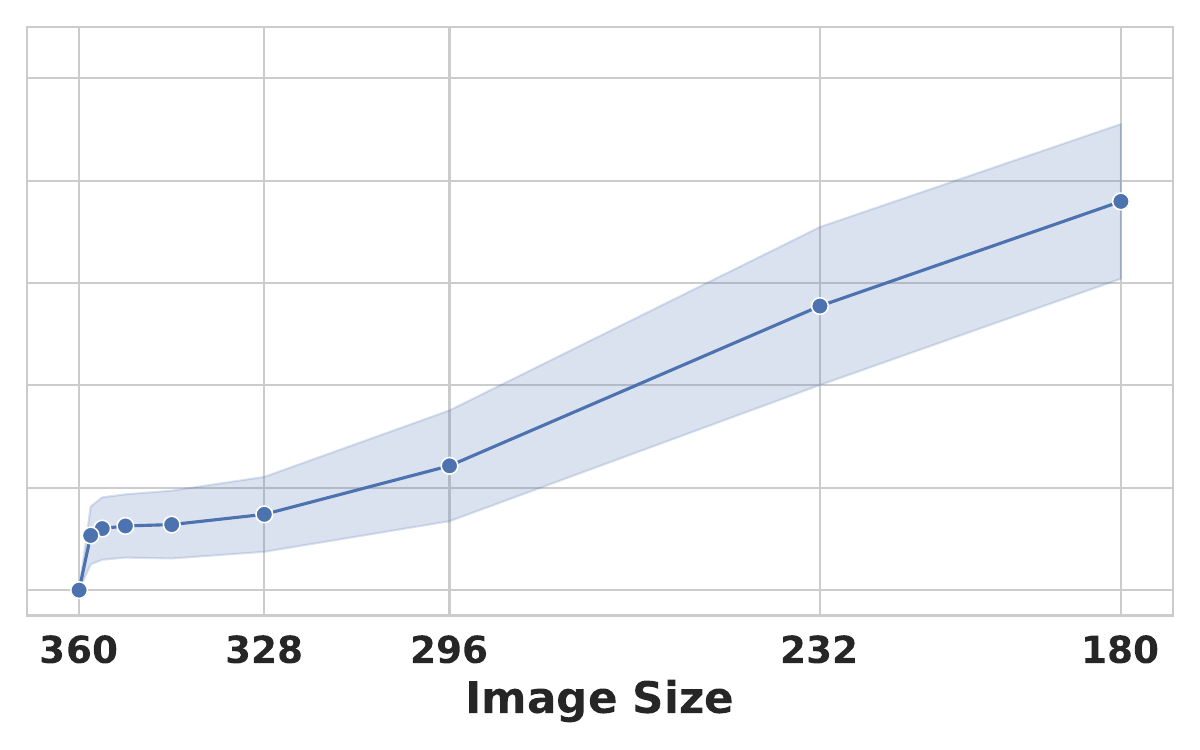}
        \caption{Downsizing}
        \label{fig:downsizing_robustness}
    \end{subfigure}
    \begin{subfigure}[b]{0.325\textwidth}
        \includegraphics[width=\linewidth]{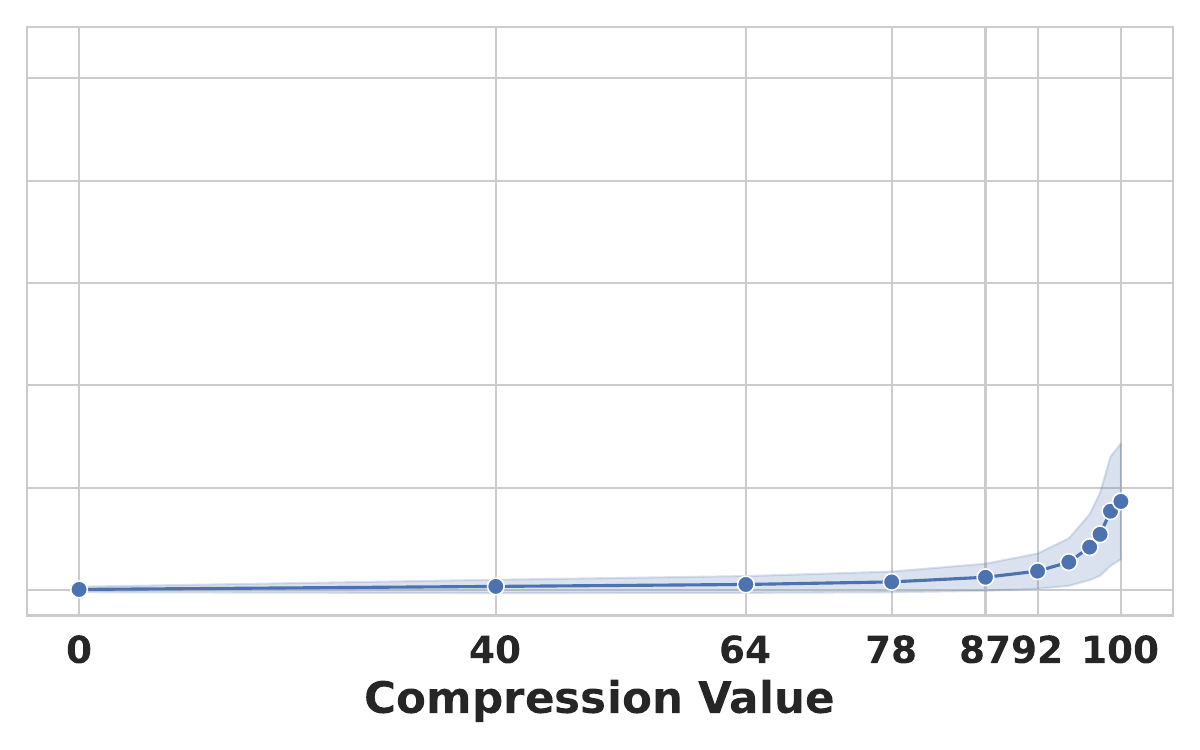}
        \caption{JPEG Compression}
        \label{fig:compression_robustness}
    \end{subfigure}
    \caption{Robustness results for image transformations computed on all ImageNet training samples (Adversary 3). The plots state the mean results and their standard deviations. The mean Hamming distance could also be interpreted as the share of hash bits expected to change.
    }
\label{fig:transformation_results}

\end{figure*}

\section{Adversary 3 -- Gradient-Free Evasion \mbox{Attacks}}\label{sec:adversary_3}
We now extend our analysis from the previous section and investigate the robustness of NeuralHash against basic, gradient-free transformations as provided by standard image editors. The more robust the hashing algorithm is, the less susceptible it is to image transformations, such as rotations, translations, or color changes. An example for each image transformation we investigated can be seen in Appendix~B.4. Many of these operations, e.g., flipping and color changes, can be reverted. This makes them particularly interesting to investigate since these transformations could be used to evade detection systems by first applying a transformation to bypass the system and then reverting them to reconstruct the original image. Unlike gradient-based attacks, the adversary does not need to have direct access to the hash algorithm or any expertise in computer science. Using a simple image editor is sufficient to perform the evasion attacks.

\begin{figure}[t!]
\centering
\includegraphics[width=.9\linewidth]{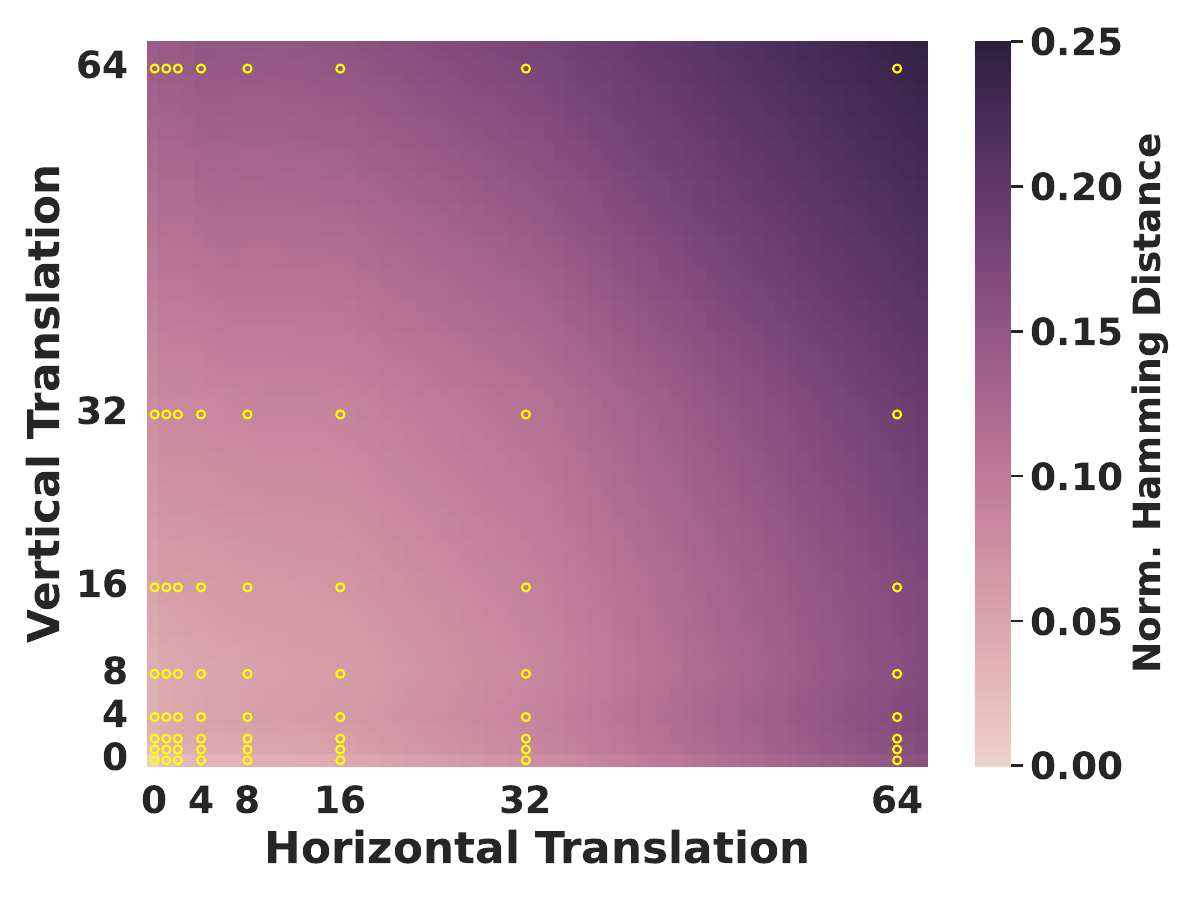}
\caption{Mean Hamming distance for all ImageNet training images translated horizontally and vertically by different values (Adversary 3). The yellow dots represent the distances for varying translations in both directions.
\label{fig:translation_robustness}}
\end{figure}

\textbf{Technical Realization.} We applied different image transformations $T(x): \mathbb{R}^{H \times W \times C} \to \mathbb{R}^{H \times W \times C}$ independently to the input images $x$. We then obtained the perceptual hash of each transformed image $T(x)$ and calculated the Hamming distance $\delta \left( H(T(x)), H(x) \right)$ between the hashes of the transformed and the original images.

For transformations with additional hyperparameters, such as degrees in rotation, we computed results for varying parameter values to take the transformations' strength into account. All investigated transformations kept the image size and removed image parts were filled with black color.

\textbf{Experimental Setup}; see Appendix~\ref{app:adv3_details} for additional details. We evaluated the robustness of NeuralHash on the 1,281,167 images from the ImageNet training split. To save computing resources, we varied most hyperparameters with an exponentially increasing step size. We investigated the following transformations independently: translation, rotation, center cropping, downsizing, flipping, changes in the HSV color space, contrast changes, and JPEG compression.

\textbf{Results.} We plot the effects of various image transformations on the Hamming distance in Figures~\ref{fig:transformation_results} and~\ref{fig:translation_robustness}. Additional plots for the other image transformations are provided in Appendix~\ref{app:hash_robustness_plots}. Our experimental results show that NeuralHash is susceptible to most, but not all, image transformations. This non-robust behavior leads to significant hash changes, even if the content of an image was only slightly modified from a human perspective. We now discuss the average effects of the different transformations investigated.

As stated in Figure~\ref{fig:translation_robustness}, image translations had a high impact on the hash calculation. Shifting an image by 32 pixels in horizontal and vertical directions resulted in a mean Hamming distance of roughly 11\%. Larger image translations even further raised the Hamming distance. Image rotations also significantly increased the Hamming distances with rotations of only one degree already changing more than 5\% of the hash bits, and rotating it by 32 degrees flipped even more than 30\% of the bits.

Similar non-robust behavior could be observed for center cropping and downsizing, where larger cropping windows and smaller image sizes increased the Hamming distance almost linearly. Furthermore, flipping the images, one of the simplest transformations without any loss of information and easy to revert, changed 29\% of the hash bits for horizontally flipping and even 37\% for vertical flipping the images.

While manipulating the hue of the images flipped 10\% to 15\% percent of the bits, varying the saturation had only minor effects on the hash computation and led to a hamming distance between 1\% and 2\%. Doubling the contrast and the brightness flipped roughly 37\% and 22\% of the hash bits. However, decreasing the values had almost no effect.

Investigating the effects of compression is especially interesting since most online services and messaging apps compress images during transmission. As can be seen in Figure~\ref{fig:compression_robustness}, compression had only a slight impact on the hashes when choosing moderate compression values. Compressing the images with a compression value of 92, which still resulted in reasonable compression and image quality, changed only roughly 2\% of the bits.

In summary, our analysis demonstrated that NeuralHash is not robust against many basic image transformations. While the system showed stronger robustness against HSV space modifications and image compression, it is more susceptible to other transformations, such as rotations or simple image flipping. The most significant hash changes occurred when image information was lost or added, e.g., by cutting parts out or adding black borders. We assume that the embedding network has been trained in a contrastive learning setting, mainly based on color transformations and, therefore, shows larger robustness against these manipulations. However, an adversary could evade detection with small effort and little to low image quality loss. In reality, the reliability of such a system is therefore questionable.

\section{Adversary 4 -- Hash Information Extraction}\label{sec:adversary_4}
In our last adversarial setting, we want to investigate whether a hash value leaks information about its corresponding image. Since the embedding network is trained with contrastive learning, we expect that the extracted feature vector contains specific information about the image. Considering these features are to some degree included in the hash value, we assume that the hash value leaks at least approximate information on an image's content. Should this assumption turn out to be true, then NeuralHash and presumably also other perceptual hash-based systems would not completely anonymize user data and still leak details about the contents on user devices. This might not be critical for a single hash value due to prediction uncertainty, but in the case of larger hash sets from the same user, an adversary might expose the general content of the images stored on the device.

\indent\textbf{Technical Realization.} To verify our assumption, we took a classification task dataset and trained a classifier to infer an image's class label given only its hash. For this, we first computed the hashes of all samples in the dataset and trained a simple classifier that takes a 96-bit vector as input.

\textbf{Experimental Setup}; see Appendix~\ref{app:adv4_details} for additional details.
We trained fully-connected neural networks with three hidden layers on hashes of the ImageNet train split with 1,000 classes and standard cross-entropy loss. To create a balanced dataset, we randomly picked 732 samples from each class resulting in a total of 732,000 training samples. We first computed the binary hashes for the samples using NeuralHash and used 90\% of the hashes from each class for training and the remaining hashes for hyperparameter optimization.

Since many of the ImageNet classes are quite similar, e.g., different dog breeds, we also used 85 coarse ImageNet categories~\cite{imagenet_categories} for training. We balanced the categories by randomly picking 980 samples from each category, resulting in 83,300 training samples, and again used 10\% of it for hyperparameter optimization. In the following, we will refer to this task as categorization.

We evaluated the networks on the official ImageNet validation split containing 50,000 samples. For evaluating the categorization model, we randomly sampled 50 samples from each category from the validation split. As evaluation metrics, we computed the top-1, top-5, and top-10 prediction accuracy's mean and standard deviation on ten different training runs and dataset splits.
\begin{table}[t]
    \centering
    \resizebox{\columnwidth}{!}{
    \begin{tabular}{lcc}
    \toprule
                                & \textbf{Classification}     & \textbf{Categorization}   \\
    \textbf{Top-1}              & $04.34\%\pm0.046\%$                      & $08.76\%\pm0.237\%$                 \\
    \textbf{Top-5}              & $12.03\%\pm0.090\%$                      & $25.85\%\pm0.423\%$                 \\
    \textbf{Top-10}             & $17.75\%\pm0.182\%$                      & $38.59\%\pm0.728\%$                 \\
    \bottomrule
    \end{tabular}
    }
  \caption{Top-1, top-5, and top-10 accuracy mean and standard deviation values computed on ten different training runs and dataset splits. Random guessing leads to an expected top-1 accuracy of only $0.1\%$ and $1.2\%$, respectively.}
  \label{tab:hash_classifier_results}
\end{table}
\begin{table}[t]
    \centering
    \resizebox{\columnwidth}{!}{
        \begin{tabular}{llll}
            \toprule
            \multicolumn{2}{c}{\textbf{Classification}}     &   \multicolumn{2}{c}{\textbf{Categorization}}   \\
            \cmidrule(lr){1-2} \cmidrule(lr){3-4}
            \textit{Class}      &   \textit{Pre.}           &   \textit{Category}   &   \textit{Pre.}           \\
            Website             &   $52.0\%$                &   Mosquito Net        &   $52.0\%$                \\
            Panda               &   $36.0\%$                &   Sloth               &   $30.0\%$                \\
            Leonberg (dog breed)           &   $36.0\%$                &   Hay                 &   $30.0\%$                \\
            \bottomrule
        \end{tabular}
    }
  \caption{Share of correctly classified hashes per class and category for one model of each task. The percentage value describes how many of the examples within a specific class or category were correctly classified. See \cref{app:adv4_add_results} for additional results.}
  \label{tab:hash_classifier_results_per_class}
\end{table}

\indent\textbf{Results.} We state our evaluation results in Table~\ref{tab:hash_classifier_results}. The results confirm our assumption that hash values indeed leak information about the corresponding images' contents. Our classifier trained on 1,000 classes achieved a top-1 test accuracy of 4.34\%. Note that ImageNet is a quite challenging dataset and random guessing only achieves an expected accuracy of 0.1\%. For the top-5 accuracy, the neural network reached an accuracy as high as 12.03\%, and even 17.75\% for the top-10 accuracy. The categorization model trained on the more coarse 85 categories predicted the correct category in 8.76\% of the cases. Note that in this case prediction through random guessing would be $1.18\%$. For the top-5 and top-10 accuracy, the model even achieved 25.85\% and 38.59\% correct predictions. The standard deviation is quite low, so random selection of training and evaluation samples has only minor effects on the performance of the models.

For further analysis, we calculated the share of correct predictions for each class and category for a randomly selected trained model and present the three classes and categories that have been predicted with the highest precision in Table~\ref{tab:hash_classifier_results_per_class}. For the classifier, images of \textit{Websites} were predicted with a precision of 52\%. For the coarse categories, \textit{Mosquito Nets} achieved a precision of 52\%. We state extended results in Appendix~\ref{app:adv4_add_results}.

In conclusion, our simple ImageNet classifiers have shown that hashes still contain information about the images' contents. And even though the hashes contain less information than the corresponding images, they should be handled with care in practical applications. An attacker with access to a user's image hashes might infer private information about contents on the device.

\section{Lessons \& Implications}\label{sec:lessons}
We will conclude our work with a summary of the lessons and their implications we draw from our experimental evaluation. \newline

\noindent
\textbf{Current systems are not robust.} Our experimental results illustrate that perceptual hashing systems like NeuralHash are in no way robust against gradient-free detection evasion attacks. Simple image modifications, such as flipping an image horizontally, already allow an attacker to evade detection in most cases, even without direct access to the system. The attacker does not need any technical knowledge and can manipulate the images with standard image editors to evade detection. %With gradient-free detection evasion attacks, the manipulations are clearly visible, e.g., black areas in the manipulated images.
Even though we focused in our experiments on NeuralHash, we are convinced that other systems are also vulnerable to this kind of attack since deep neural networks react sensitively to input changes, as also adversarial attacks illustrate. As we outline later, the networks might be trained to be more robust against simple image manipulations but will never be entirely secure. A more robust system will force the attacker to alter the images stronger, but the content of the images will most likely still be visible and recognizable. For example, disassembling an image into smaller parts for storing and reassembling the image only for viewing would lead to completely different hashes for each part without any information loss in the full image.

On the other hand, evasion attacks exploiting gradient information allow evasion with only minor image manipulations. Our experiments have shown that with a bit of technical knowledge, an attacker can modify the images to evade detection without the modifications being clearly visible. Such an attack will most likely be always possible in a client-side scanning system since the attacker will always have access to the model on the user device and is, therefore, able to calculate the gradient of the model. \newline

\noindent
\textbf{Client-side scanning systems can be misused for malicious purposes.} Proponents of client-side scanning emphasize the fact that the systems enable the detection of CSAM and other criminal material while avoiding backdoor keys in the end-to-end encryption and maintaining the user's privacy. However, client-side scanning opens the door for other malicious attacks. While evasion of the detection system only renders the system useless, an attacker can further misuse the system for framing or monitoring innocent users using hash collision attacks. As we have demonstrated, an attacker can slightly alter images to change their hashes to a specific value, causing false-positive detections for arbitrary image contents. With access to an official hash database or a surrogate, an attacker can frame innocent people by sending these manipulated images to their devices. As a result, without even knowing, the receivers of such images get flagged.

As legislators call for preventing the distribution of harmful material while maintaining encryption gets louder, client-side scanning systems using perceptual hashing look very promising at first glance. However, governments or organizations with control over the client-side scanning system could share manipulated images in social media, and users that downloaded a subset of these images get flagged. Mass surveillance of people with undesired beliefs and opinions is therefore simply and secretly realizable. Additional tools or backdoors in the user devices are not needed.

Also, there is no guarantee that the hash database will not be extended with additional, non-criminal content for surveillance. Since the databases are not publicly available, changes are not traceable, and the targeted content might be anything -- who controls the provider of such a system? \newline

\noindent
\textbf{Hash values still contain information.} Even though the hash values of images contain much less information than the images themselves, our experiments have shown that there is still some information encoded in these hash values. Using the hash values, inference about the general content of images on the device is possible. Although image information is highly compressed, sharing these hashes with the system's provider might harm a user's privacy. Moreover, hashes with a larger number of bits are expected to leak even more information due to an increased information capacity. Apparently, Microsoft's PhotoDNA, which is already widely used for CSAM detection, produces hashes with 144-byte values~\citep{photodna_blog}. We expect these hashes to leak even more information than NeuralHash, whose hashes only consist of 96 bits or 12 bytes. \newline

\noindent
\textbf{Mitigating the risks.} Regarding the technical aspects of the system, one way to mitigate the effectiveness of collision attacks against hashing-based detection systems is to install another server-side hashing procedure not available to the attacker. But this would, in turn, imply that the images are not encrypted on the server or could be decrypted by the provider, questioning the promised privacy advantages. Another method would be to restrict the model access and prevent gradient computations, which would indeed make gradient-based attacks hardly feasible. However, these steps do not improve the robustness against standard image transformations.

Client-side scanning methods based on neural networks will most likely always enable gradient computation or approximation and, consequently, facilitate arbitrary hash manipulations. Hence, even an updated version of NeuralHash's embedding network would very likely suffer from the demonstrated vulnerabilities.

To complicate hash manipulations, we expect training the embedding network in an adversarial setting~\cite{fgsm, Wang21} might improve its robustness against attacks that manipulate the images. In adversarial training, a sample is first manipulated by one of our gradient-based attacks or standard image transformations and then used as a regular training sample to update the model's weights.

It is further important to restrict public access to the hash database. If the plain hashes are leaked, it is even easier for an adversary to produce a large number of false-positive matches. On the other hand, an independent instance should monitor the database to avoid manipulations and the introduction of non-criminal material. \newline

\noindent
\textbf{Summary.} Our work points out that NeuralHash, and arguably deep perceptual hashing algorithms in general, are not robust and susceptible against gradient-based and gradient-free image manipulations. It is questionable if neural networks in their current form will ever be fully robust against these kind of attacks.

We further share the viewpoint that client-side scanning might have initially reasonable goals but also acknowledge the potential risks of expanding the scope of application since it is running on all devices regardless of whether crime is suspected or not. We hope that service providers and governments are not going in this direction in crime detection due to the many technical and societal risks.

From a technical and ethical viewpoint, we conclude that NeuralHash and related client-side scanning systems do not provide a safe method for detecting legal violations and should not be deployed on user devices since attackers, service providers, and governments could simply manipulate and misuse the systems according to their interests. After all, mobile devices contain a lot of sensitive information about their users, ranging from dating behavior to health care and financial status.

Instead, based on our results, we call for NeuralHash and other client-side scanning systems not to be installed in practice, due to their manipulability, risk of abuse, and lack of robustness. Granting access to the content stored on the devices, even only the images, might pave the way for further interventions in a user's privacy. We should, therefore, be very careful in deploying detection systems directly on user devices.

\section{Conclusion}\label{sec:conclusion}
Using NeuralHash as a practical example from a real-world use case, our work demonstrated that perceptual hashing-based detection systems might be cryptographically well-proven but are still highly susceptible to various attacks, some of them trivial, that crack the system.
We empirically showed that the system is vulnerable to collision attacks that might lead to serious consequences for privacy and allow monitoring of users beyond the intended use case of CSAM detection.

Our results further suggest that the feature extraction step in the hash computation is in no way robust against gradient- and transformation-based image manipulations and induces vulnerabilities that allow an attacker to easily evade detection.
Moreover, even though NeuralHash produces only short bit-string hashes, we practically demonstrated that the hashes still leak some information on the images' contents and should be treated with caution.
We conclude that NeuralHash, and deep perceptual hashing systems in general, pose several risks to the reliability of the systems and even to human society. Therefore, we believe that NeuralHash, and other client-side detection systems, should not be installed in their current state. \newline

\noindent
\textbf{Reproducibility Statement.} We deliberately decided to make our source code publicly available to reproduce the experiments and investigate existing and future perceptual hashing systems: \href{https://github.com/ml-research/Learning-to-Break-Deep-Perceptual-Hashing}{https://github.com/ml-research/Learning-to-Break-Deep-Perceptual-Hashing}. \newline

\noindent
\textbf{Acknowledgments.} This work was supported by the German Ministry of Education and Research (BMBF) within the framework program ``Research for Civil Security'' of the German Federal Government, project KISTRA (reference no. 13N15343). It also benefited from the National Research Center for Applied Cybersecurity ATHENE, a joint effort of BMBF and the Hessian Ministry of Higher Education, Research, Science and the Arts (HMWK). \newline

\noindent
\textbf{Disclaimer.} The views and opinions expressed in this article are those of the authors and do not reflect the official policy or position of the authors' institutions or any agency of the German Federal Government.

%%%%%%%%% REFERENCES
\vfill
{\small
\bibliographystyle{plainnat}
\bibliography{references}
}
\newpage
\appendix

\section{Experimental Details}
Here we state the technical details of our experiments to improve reproducibility and eliminate ambiguities.

\subsection{Hard- and Software Details}\label{app:experimental_details}
We performed all our experiments on NVIDIA DGX machines running NVIDIA DGX Server Version 4.4.0 and Ubuntu 18.04 LTS. The machines have 1.6TB of RAM and contain Tesla V100-SXM3-32GB-H GPUs and Intel Xeon Platinum 8174 CPUs. We further relied on Python 3.7.10 and PyTorch 1.9.0 with Torchvision 0.10.0 ~\cite{pytorch} for the implementation of the embedding network and image optimizations. The various StyleGAN2 models are available at \url{https://github.com/NVlabs/stylegan2-ada-pytorch}.

We manually extracted~\cite{neuralhash2onnx} the NeuralHash model, including the network weights and hash matrix, from a Mac running macOS Big Sur Version 11.6. We then rebuilt the network's architecture in PyTorch and assigned the extracted weights appropriately. This allows us to run our experiments on GPUs, significantly increasing the inference speed. Also, gradients can be computed by PyTorch's automatic differentiation.

To check if the change of framework induced a significant deviant behavior of the model, we computed hashes for all 100,000 samples from the ImageNet test split with our PyTorch model and a separately extracted ONNX model~\cite{neuralhash2onnx}. Only for a single bit in one of the hashes, both models differed, demonstrating that both models are virtually identical and our results should be transferable to the original NeuralHash system.

\subsection{Experimental Details for Adversary 1}\label{app:adv1_details}
For our analyses, we created a surrogate CSAM hash database by hashing all samples from the Stanford Dogs dataset~\cite{dogs_dataset}. The dataset contains 20,580 high-quality images of 120 dog breeds, so the images share similar content. We then performed our attacks by modifying the samples from the ImageNet ILSVRC2012~\cite{deng2009imagenet, ILSVRC15} test split containing 100,000 images of 1,000 different classes, from which we took the first 10,000 samples. Note that the Stanford Dogs samples are taken from the ImageNet dataset but only have a total overlap with the test split of 18 images, 2 of them in the first 10,000 samples. We removed these images from the results to avoid any evaluation biases. For each ImageNet sample $x$, we then first computed its hash and looked up the hash $\hat{H}$ with the smallest Hamming distance in the database.

We used the Adam optimizer~\cite{adam} to directly optimize $x$ and set the learning rate to $10^{-3}$ and $\beta$ to $(0.9, 0.999)$. We further set $\lambda=100$ and stopped the optimization when either $H(x)=\hat{H}$ was satisfied or aborted after 10,000 iterations. If we do not stop the optimization process when $H(x)=\hat{H}$ is reached, the image quality might be further improved due to the SSIM term.

For the GAN-based attack, we used Adam with a learning rate of $10^{-2}$ and $\beta=(0.1, 0.1)$ to optimize the style vectors. We used StyleGAN2 models pretrained on FFHQ~\cite{Karras19} (faces), MetFaces~\cite{Karras2020ada} (faces in Art), AFHQ Dogs~\cite{choi2020starganv2} (dogs), and BreCaHAD~\cite{BreCaHAD} (breast cancer histopathology).

\subsection{Experimental Details for Adversary 2}\label{app:adv2_details}
We performed our evasion attacks on the first 10,000 samples of the ImageNet ILSVRC2012 test split. We used the Adam optimizer to directly optimize an image $x$ and set the learning rate to $10^{-3}$ and $\beta$ to $(0.9, 0.999)$ for the Standard and Edges-Only attacks. We applied a decaying weight for the SSIM term and set $\lambda=5\cdot 0.99^{step}$ where $step$ denotes the number of optimization steps already performed. By decaying the weight, we avoided that the optimizer gets stuck in local minima. We set the standard deviation of the $3\times 3$ Gaussian filter in the Canny algorithm to $3$ and the hysteresis thresholds to $0.1$ and $0.2$ (value range of the greyscale images is $[0, 1]$), respectively. For both attacks, we performed a maximum of 1,000 optimization steps.

In the Few-Pixels attack, we set the learning rate to $1.0$ and kept $\beta$ the same as before. We did not apply the SSIM term to ease the optimization and, consequently, set $\lambda=0$. We also applied a learning rate scheduler and halved the learning rate after five and ten optimization steps. The scheduler and optimizer are re-initialized each time new pixels were added to the set. We aborted the optimization after 150 pixels without a hash change. In all three attacks, we set the minimum Hamming distance to $\delta_0=0$ and, consequently, stopped each attack when $H(x) \neq \tilde{H}$.

Assuming that the hash database also contains the Hamming neighbors of each hash, we can force our attacks to produce larger perturbations and move a sample even further away in terms of the Hamming distance by setting $\delta_0>0$. We repeated our Standard attack with minimal Hamming distances $\delta_0 \in [0, 0.5]$ and increased $\delta_0$ in steps of $0.05$.

\subsection{Experimental Details for Adversary 3}\label{app:adv3_details}
We used the transformation package of Torchvision~\cite{pytorch} to perform all transformations and additionally used imgaug~\cite{imgaug} to investigate the effect of JPEG compression.

\subsection{Experimental Details for Adversary 4}\label{app:adv4_details}
The fully-connected neural network consists of three hidden layers. After each layer, we added a batch normalization layer and also applied dropout regularization to prevent overfitting. While the first and the last hidden layers consist of 2,048 neurons each, the middle layer contains 4,096 neurons. We trained the classifiers using the Adam optimizer with a learning rate of $10^{-3}$ and set $\beta$ to (0.9, 0.999). For the classification network, a batch size of $64$ was used. For training the categorization network, we used a batch size of $128$ and a weight decay value of $10^{-3}$. We stopped the training of all networks when the loss on the custom validation set did not decrease by at least $10^{-4}$ for $10$ epochs.
We set the dropout probability to $0.3$ for the classification and $0.2$ for the categorization task.

For the categorization task, we removed the category \textit{other} and created for each of the classes in this category a separate category. This results in a total of 85 categories. We balanced the categories by randomly picking 980 samples from each category, resulting in 83,300 training samples, and used stratified sampling to create a custom validation set consisting of 10\% of the training set.

\section{Additional Experiments and Results}
In this section, we state additional experiments and results that did not fit in the main part of the paper.

We also show results for our GAN-based preimage attack in \cref{fig:gan_hash_collisions}. We used StyleGAN2 models pretrained on MetFaces, FFHQ, AFHQ dogs, and BreCaHAD.

\subsection{Adversary 1 -- Hash Collision Examples}\label{app:add_hash_collision_samples}
\begin{figure*}
\centering
\includegraphics[width=\linewidth]{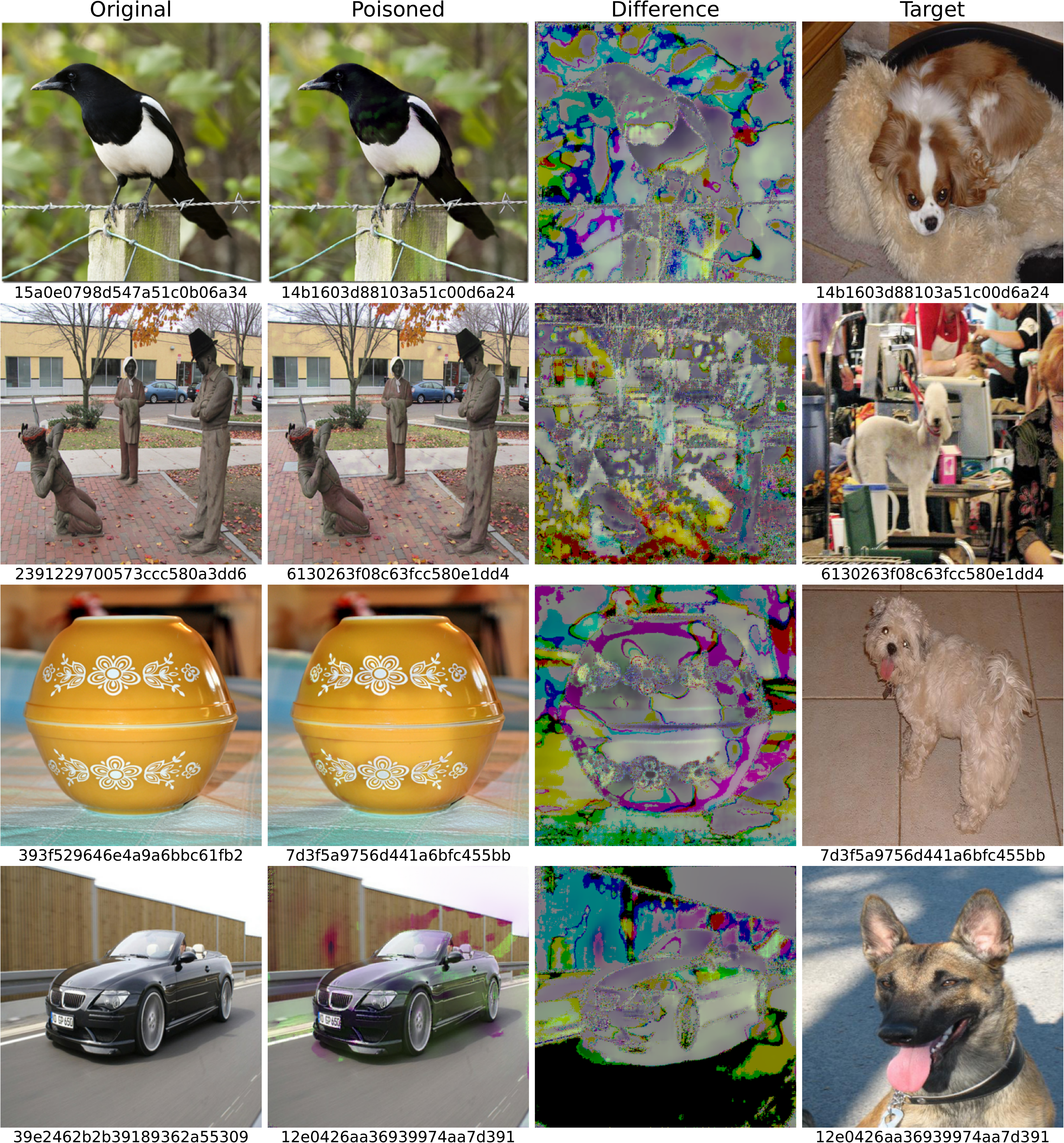}
\caption{ Selected samples of our hash collision attack. We modify an original image to have the same hash as a target sample from our surrogate hash database. We state the computed hash below each image. We further visualize the differences between the modified and original images. The differences are normalized to $[0, 1]$-space to make them more visually apparent. Black marks areas where no change was made.}
\label{fig:add_hash_collision_samples}
\end{figure*}

We state further examples of our hash collision attacks from \cref{sec:adversary_1} in \cref{fig:add_hash_collision_samples}. It illustrates the image manipulations of four different images, together with the added image distortions and the image corresponding with the target hash.

\subsection{Adversary 1 -- GAN-Based Approach}\label{app:adv1_gan}
We propose a second approach to achieve the goals of our Adversary 1 from \cref{sec:adversary_1}. It is a basic preimage attack and uses a generative adversarial network (GAN)~\cite{gan_goodfellow} to generate an image that yields an arbitrarily predefined hash. We used various pretrained StyleGAN2~\cite{Karras2019stylegan2, Karras2020ada} models to create synthetic images that produce a target hash. We first sampled a random Gaussian noise vector, which we mapped into a style vector representation by StyleGAN's mapping network. The style vector acts as input for the generator, which then creates a corresponding image. We resized the generated image to $360\times 360$ and computed the Hinge loss, as stated in Eq.~\ref{eq:hinge}. We then propagated the loss back through the generator and directly optimized the style vector to force the generated image toward the target hash. We used Adam to directly optimize the style vectors. We used StyleGAN2 models pretrained on FFHQ~\cite{Karras19} (faces), MetFaces~\cite{Karras2020ada} (faces in Art), AFHQ Dogs~\cite{choi2020starganv2} (dogs), and BreCaHAD~\cite{BreCaHAD} (breast cancer histopathology).

\begin{figure*}
\centering
\includegraphics[width=\linewidth]{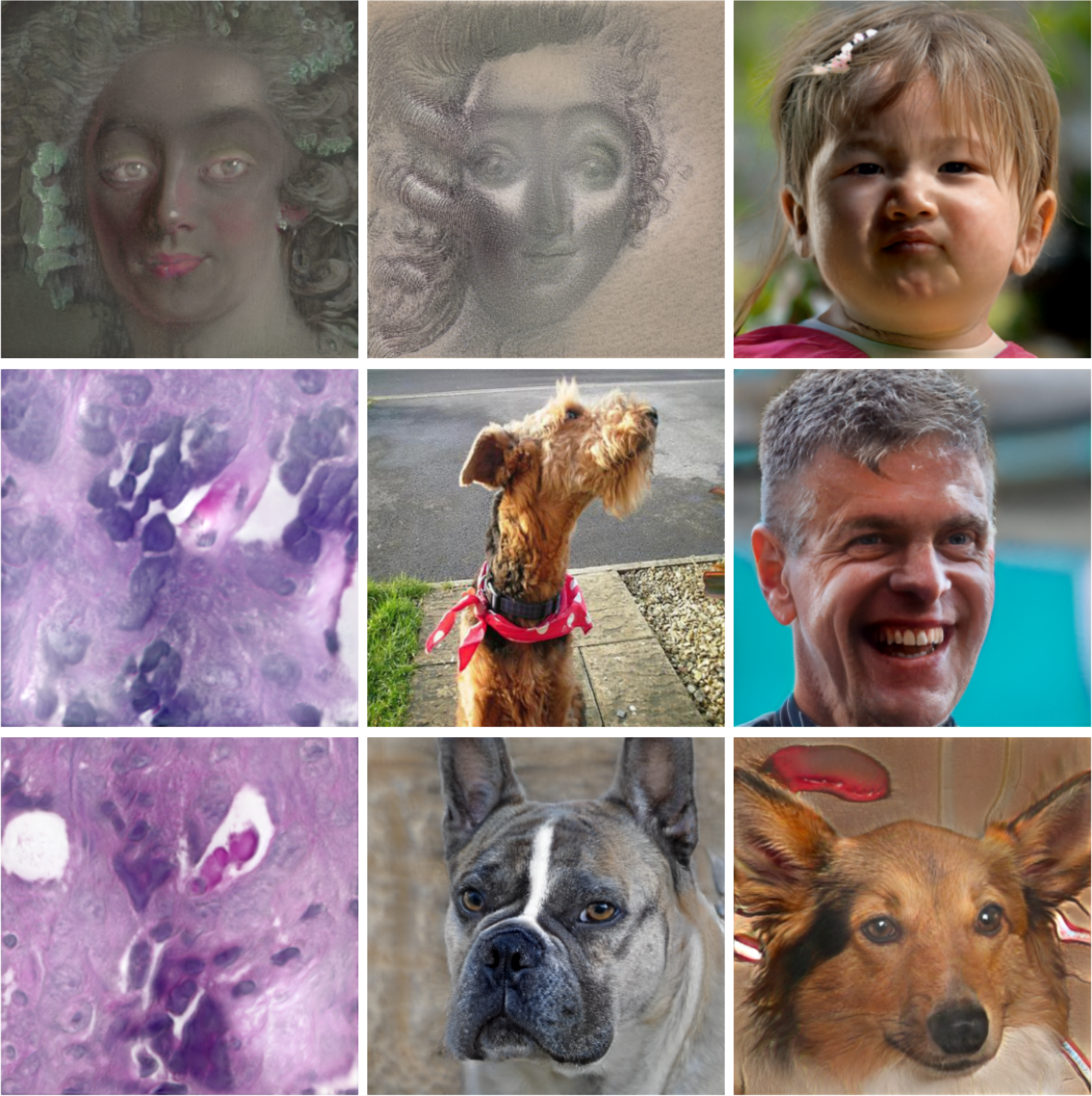}
\caption{ Samples of our StyleGAN-based collision attack. We directly optimized the style vectors to generate images approaching the target hash (center image). All images produce an identical hash value. We used StyleGAN2 models trained on MetFaces (top left), FFHQ (top right), AFHQ dogs (bottom right) and BreCaHAD (bottom left).}
\label{fig:gan_hash_collisions}
\end{figure*}

We were able to create various visually different images that all produce the exact same hash value. We plot some of the samples in Figure~\ref{fig:gan_hash_collisions}. We point out that the generated images are not free from artifacts and can easily be recognized as fake images. Still, our results open up the question of whether a GAN could be trained to generate realistic images for a given hash value. We leave building such a model for future work.

\subsection{Adversary 2 -- Hash Change Examples}\label{app:add_hash_change_samples}
\begin{figure*}
\centering
\includegraphics[width=\linewidth]{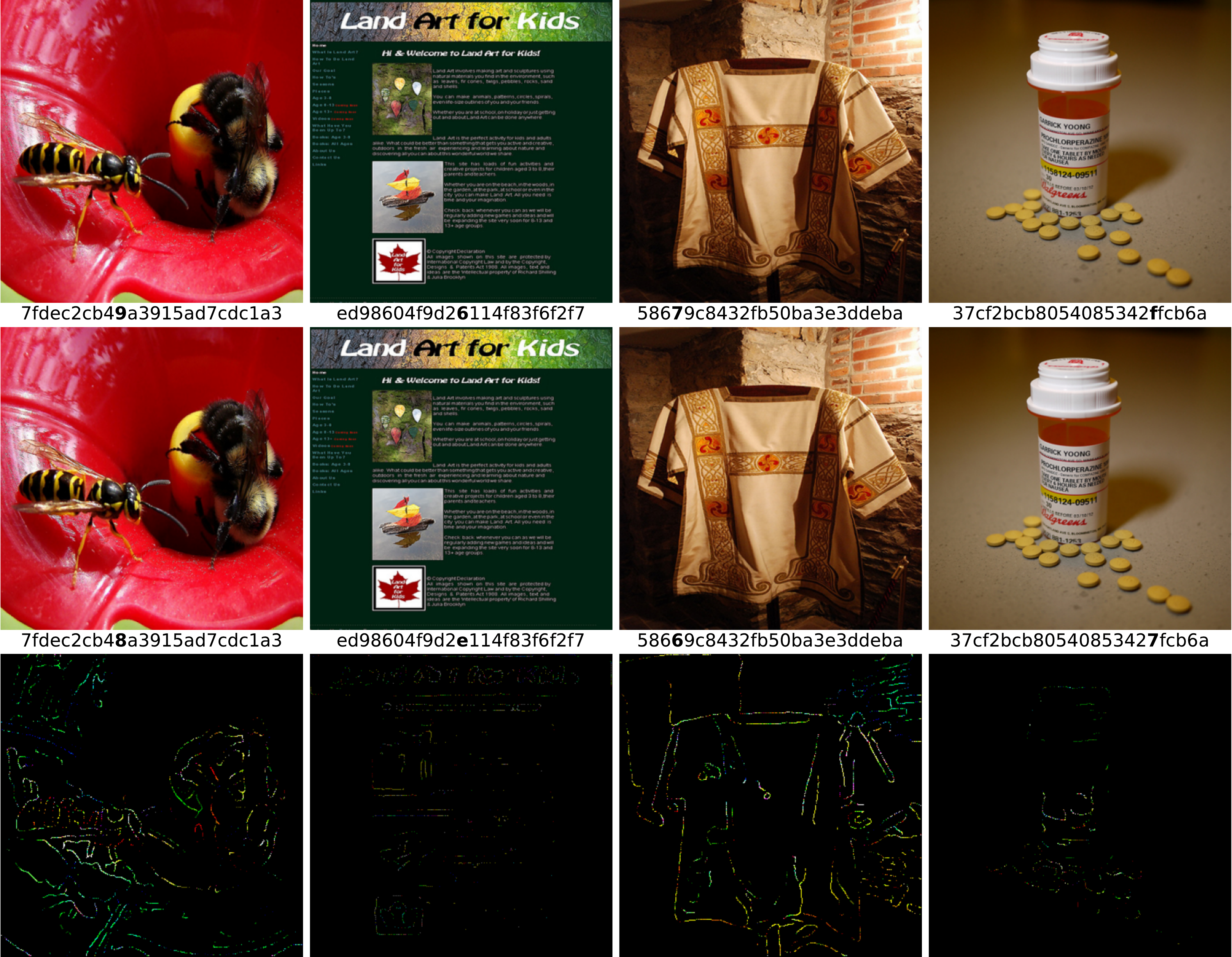}
\caption{ Selected samples of our Edges-Only evasion attack where we only modify pixels that belong to an edge. We state the computed hash below each image and highlight differences. We further visualize the differences between the modified and original images. The differences are normalized to $[0, 1]$-space to make them more visually apparent. Black marks areas where no change was made.}
\label{fig:add_hash_change_edges}
\end{figure*}

\begin{figure*}
\centering
\includegraphics[width=\linewidth]{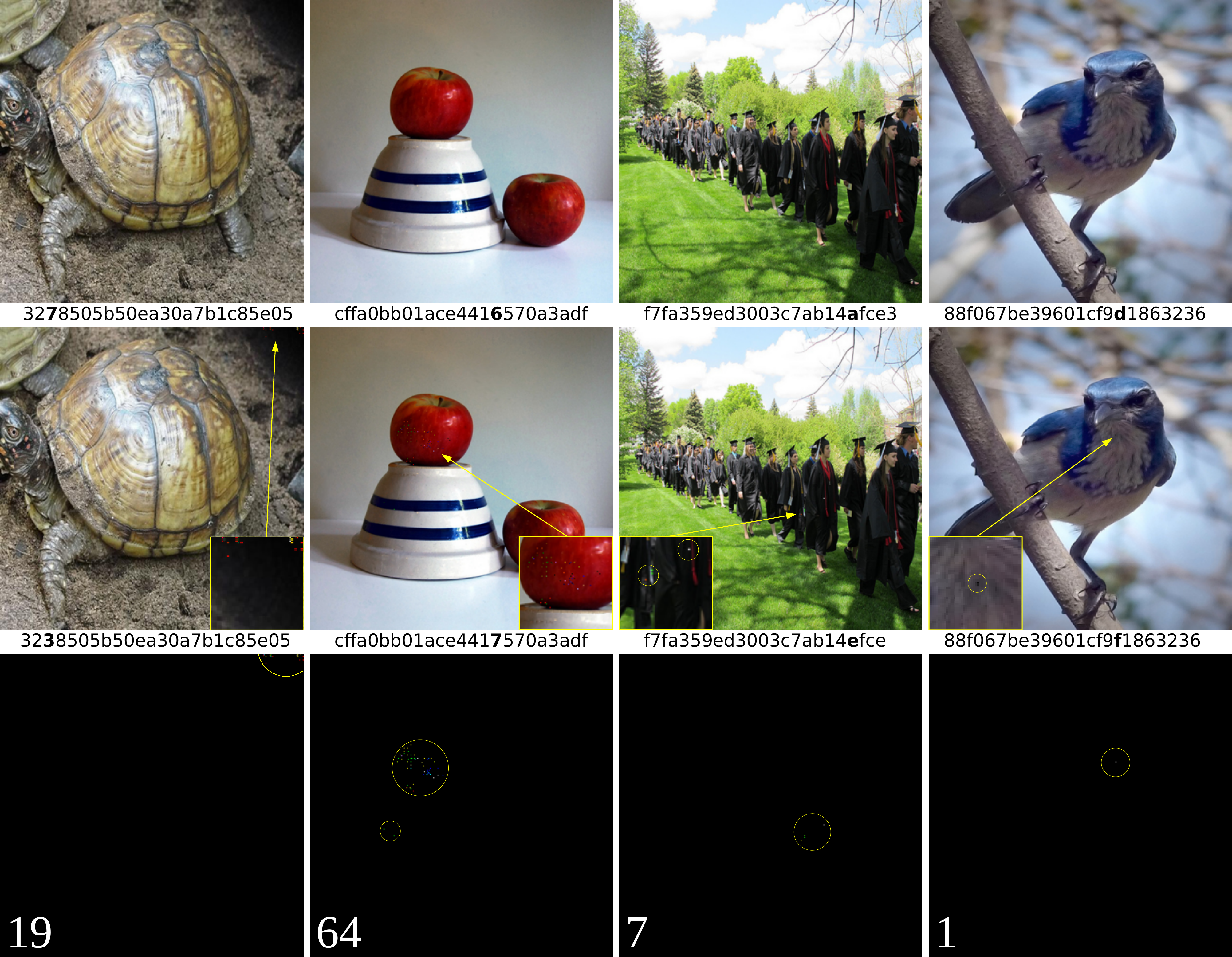}
\caption{ Selected samples of our Few-Pixels evasion attack where we try to change as few pixels as possible. We state the computed hash below each image and highlight differences. We further visualize the differences between the modified and original images. The differences are scaled by a factor of 10 to make them more visually apparent. Black marks areas where no change was made.}
\label{fig:add_hash_change_pixels}
\end{figure*}

\begin{figure*}
\centering
\includegraphics[width=0.85\linewidth]{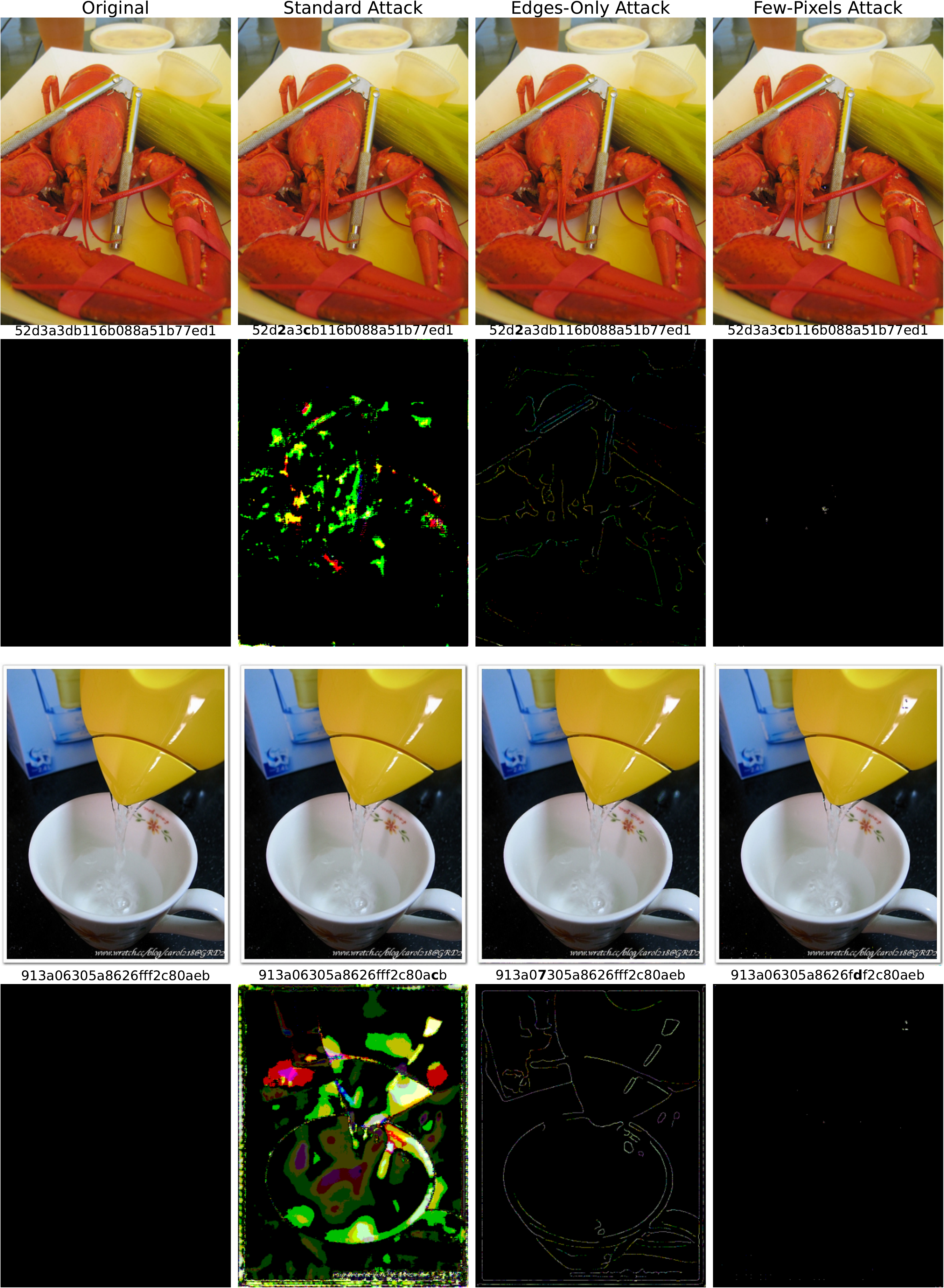}
\caption{ Visualization of our gradient-based evasion attacks against original-sized inputs. We make use of the differentiability of resizing operations by linear interpolation, allowing us to propagate gradients back to the original image.}
\label{fig:add_hash_change_no_resize}
\end{figure*}

We also demonstrate additional results to our gradient-based evasion attacks from our Adversary 2 in \cref{sec:adversary_2}. \cref{fig:add_hash_change_pixels} visualizes the original and modified images together with the induced changes for our Standard attack. Figures~\ref{fig:add_hash_change_edges} and \ref{fig:add_hash_change_pixels} show further examples for our Edges-Only, and Few-Pixels attacks. Our approaches are also applicable to the original images before resizing them. To demonstrate this fact, we state examples of all three attacks in \cref{fig:add_hash_change_no_resize}.

\subsection{Adversary 3 -- Visualizations of Transformations}\label{app:transformations}
We visualize the effects of the various transformations investigated with our Adversary 3 in \cref{sec:adversary_4}. We applied each transformation independently of each other to avoid overlapping effects on the hash computation. We assume that by combining different transformations, the effects on the hashes even further increase.

\subsection{Adversary 3 -- Transformation Results}\label{app:hash_robustness_plots}
\begin{figure*}
    \centering
    \captionsetup{justification=centering}
    \begin{subfigure}[t]{.4\textwidth}
        \centering
        \includegraphics[width=\textwidth]{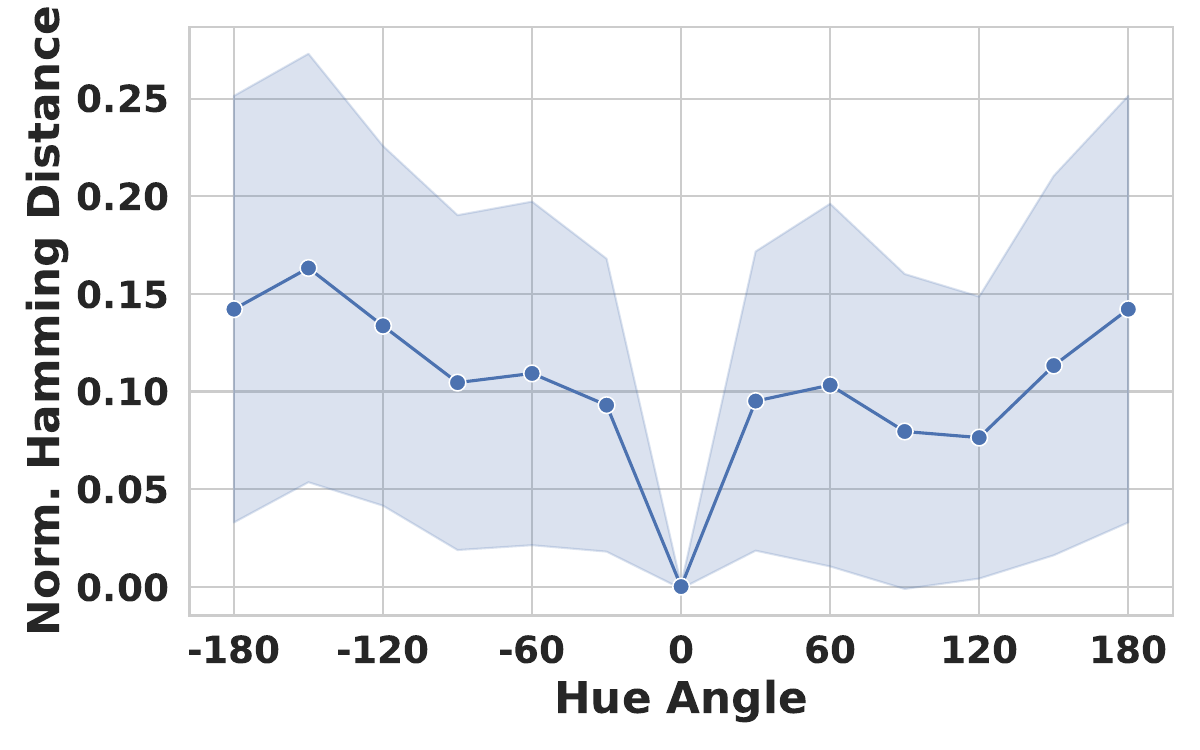}
        \caption{Hue Changes}
        \label{fig:results_hue}
    \end{subfigure}
    \begin{subfigure}[t]{.4\textwidth}
        \centering
        \includegraphics[width=\textwidth]{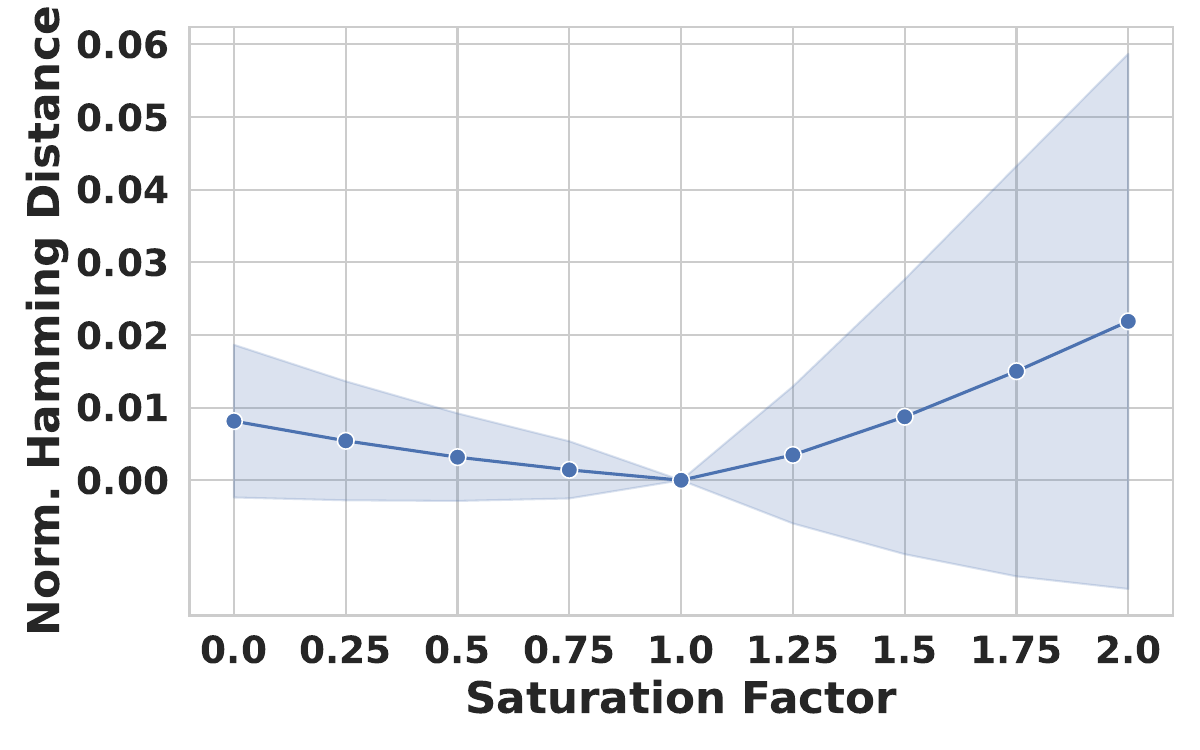}
        \caption{Saturation Changes}
        \label{fig:results_saturation}
    \end{subfigure}
    \caption{Additional plots visualizing the mean Hamming distances and their standard deviations for different gradient-free image transformations.}
    \label{fig:add_transformation_results}
\end{figure*}

In addition to our experimental results for Adversary 3 in \cref{sec:adversary_3}, we state further results for gradient-free image transformations and varying parameters in \cref{fig:add_transformation_results}. These include results for changes in hue (\ref{fig:results_hue}) and saturation (\ref{fig:results_saturation}). The Hamming distance specifies the percentage of hash bits flipped on average. All values are computed on the whole ImageNet training split.

\subsection{Adversary 4 -- Additional Results}\label{app:adv4_add_results}
For Adversary 4 in~\cref{sec:adversary_4}, we calculated the share of correct predictions for each class and category for a randomly selected trained model and present the three classes and categories that have been predicted with highest precision in ~\cref{tab:add_hash_classifier_results_per_class}.
\begin{table}[t!]
    \centering
    \begin{tabular}{llll}
        \toprule
        \multicolumn{2}{c}{\textbf{Classification}}     &   \multicolumn{2}{c}{\textbf{Categorization}}   \\
        \cmidrule(lr){1-2} \cmidrule(lr){3-4}
        \textit{Class}      &   \textit{Pre.}           &   \textit{Category}   &   \textit{Pre.}           \\
        Website             &   $52.0\%$                &   Mosquito Net        &   $52.0\%$                \\
        Panda               &   $36.0\%$                &   Sloth               &   $30.0\%$                \\
        Leonberg            &   $36.0\%$                &   Hay                 &   $30.0\%$                \\
        Carbonara           &   $34.0\%$                &   Shark               &   $28.0\%$                \\
        Earthstar           &   $34.0\%$                &   Coral               &   $26.0\%$                \\
        Nematode            &   $34.0\%$                &   Honeycomb           &   $24.0\%$                \\
        Odometer            &   $34.0\%$                &   Arachnid            &   $22.0\%$                \\
        Sealyham Terrier    &   $30.0\%$                &   Fungus              &   $22.0\%$                \\
        Daisy               &   $30.0\%$                &   Boat                &   $20.0\%$                \\
        Valley              &   $30.0\%$                &   Bear                &   $20.0\%$                \\
        \bottomrule
    \end{tabular}
  \caption{Share of correctly classified hashes per class and category for one model of each task. The percentage value describes how many of the examples within a specific class or category were correctly classified.}
  \label{tab:add_hash_classifier_results_per_class}
\end{table}

\clearpage
%%%%%%%%%%%%%%% Figures

\begin{figure*}
\centering
\includegraphics[width=\linewidth]{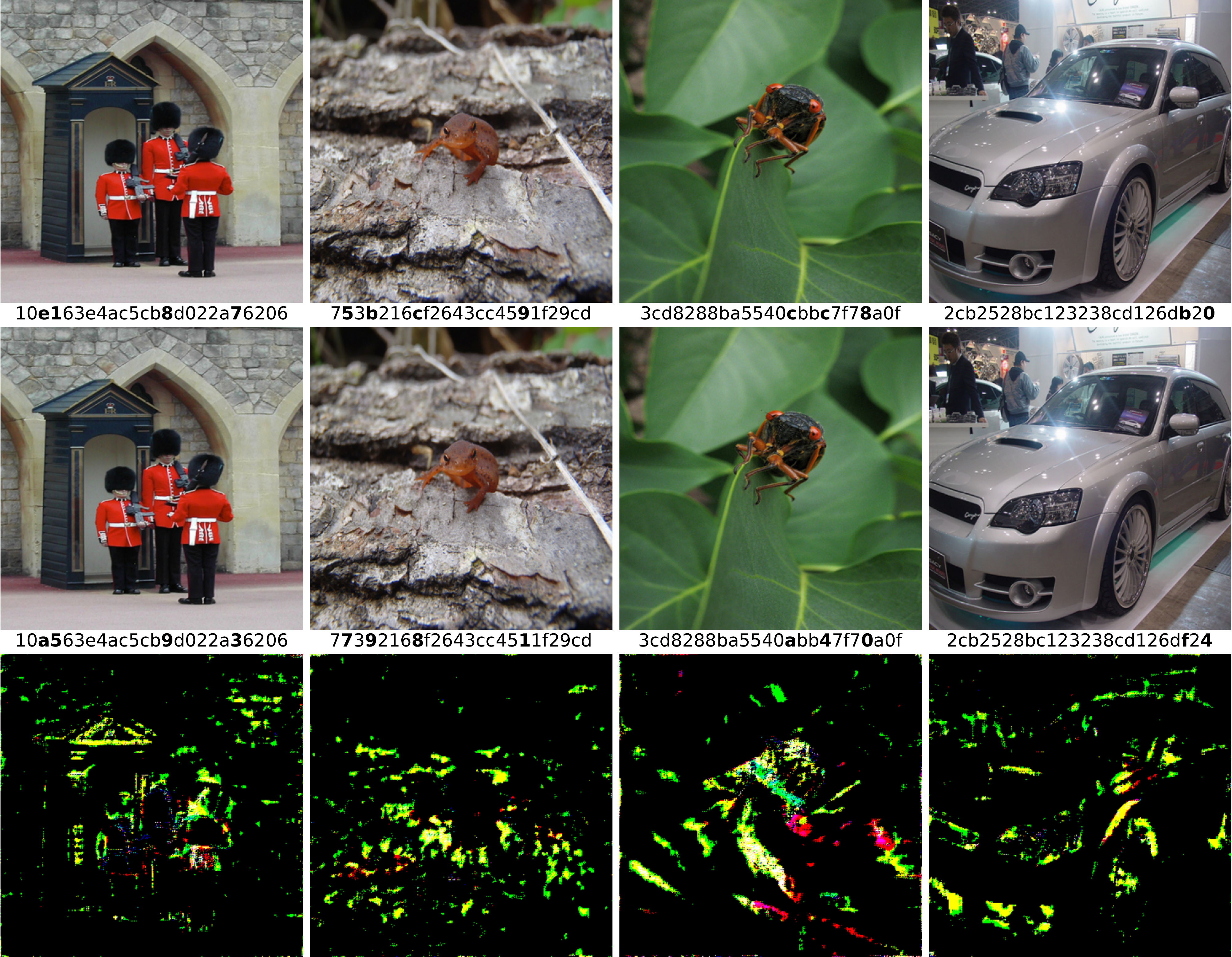}
\caption{ Selected samples of our Standard evasion attack where we allow modifications of all pixels in an image. We state the computed hash below each image and highlight differences. We further visualize the differences between the modified and original images. The differences are normalized to $[0, 1]$-space to make them more visually apparent. Black marks areas where no change was made.}
\label{fig:add_hash_change_standard}
\end{figure*}

\begin{figure*}
    \begin{center}
    \captionsetup[subfigure]{labelformat=empty}
    \captionsetup{justification=centering}
    \begin{subfigure}[t]{.16\textwidth}
        \centering
        \includegraphics[width=\textwidth]{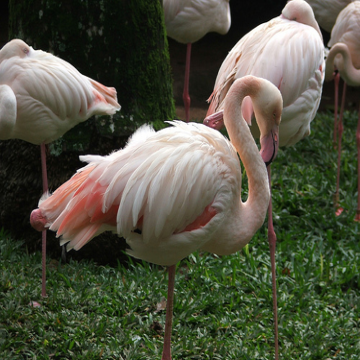}
        \caption{original}
    \end{subfigure}
    \begin{subfigure}[t]{.16\textwidth}
        \centering
        \includegraphics[width=\textwidth]{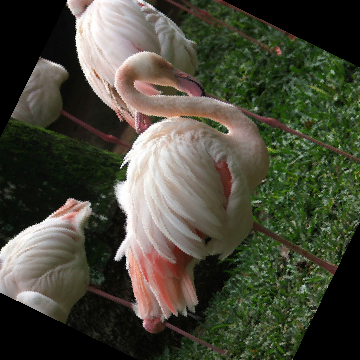}
        \caption{rotated}
    \end{subfigure}
    \begin{subfigure}[t]{.16\textwidth}
        \centering
        \includegraphics[width=\textwidth]{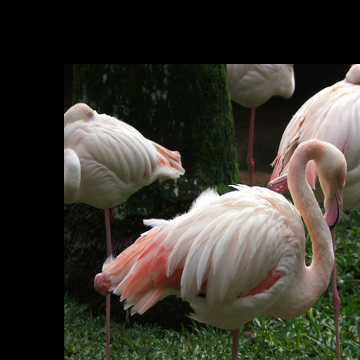}
        \caption{translated}
    \end{subfigure}
    \begin{subfigure}[t]{.16\textwidth}
        \centering
        \includegraphics[width=\textwidth]{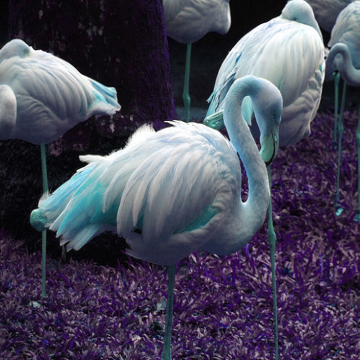}
        \caption{hue changed}
    \end{subfigure}
    \begin{subfigure}[t]{.16\textwidth}
        \centering
        \includegraphics[width=\textwidth]{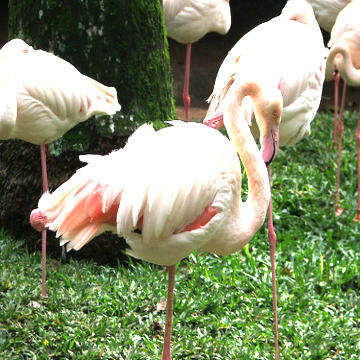}
        \caption{brightness changed}
    \end{subfigure}
    \begin{subfigure}[t]{.16\textwidth}
        \centering
        \includegraphics[width=\textwidth]{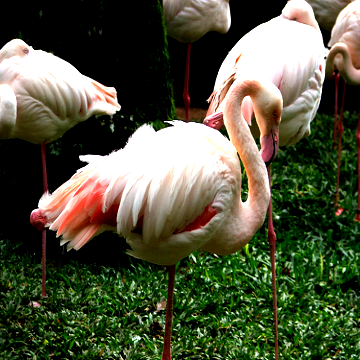}
        \caption{contrast changed}
    \end{subfigure}
    \begin{subfigure}[t]{.16\textwidth}
        \centering
        \includegraphics[width=\textwidth]{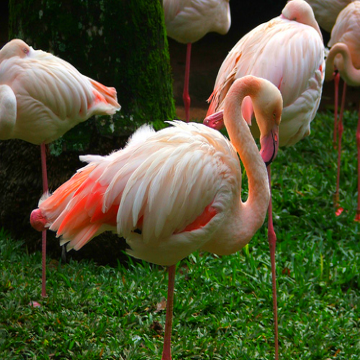}
        \caption{saturation changed}
    \end{subfigure}
    \begin{subfigure}[t]{.16\textwidth}
        \centering
        \includegraphics[width=\textwidth]{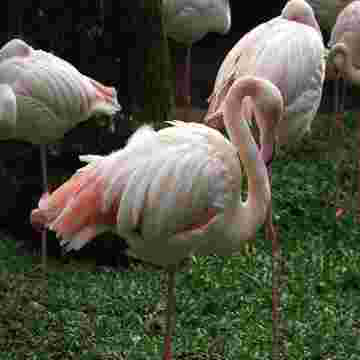}
        \caption{compressed}
    \end{subfigure}
    \begin{subfigure}[t]{.16\textwidth}
        \centering
        \includegraphics[width=\textwidth]{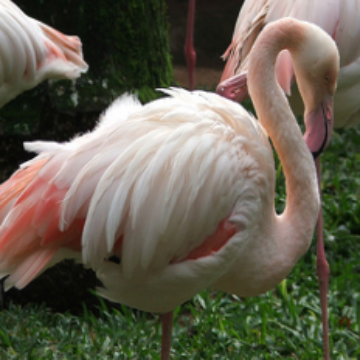}
        \caption{center cropped}
    \end{subfigure}
    \begin{subfigure}[t]{.16\textwidth}
        \centering
        \includegraphics[width=\textwidth]{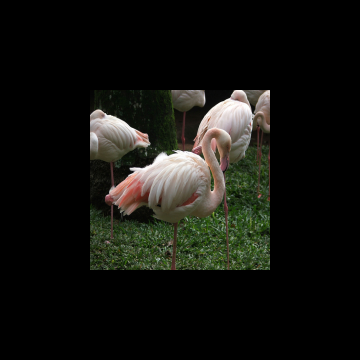}
        \caption{downsized}
    \end{subfigure}
    \begin{subfigure}[t]{.16\textwidth}
        \centering
        \includegraphics[width=\textwidth]{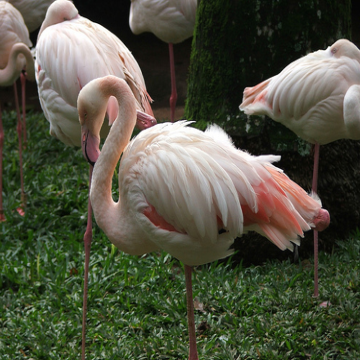}
        \caption{horizontally flipped}
    \end{subfigure}
    \begin{subfigure}[t]{.16\textwidth}
        \centering
        \includegraphics[width=\textwidth]{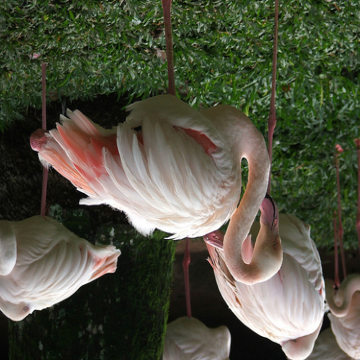}
        \caption{vertically flipped}
    \end{subfigure}
    \end{center}
\caption{ Samples of the image transformations for which we tested the robustness of NeuralHash. Many of these operations can be reversed, which is why they could be misused to bypass perceptual detection systems without degrading the image quality by first applying a transformation and then reverting it after the detection system has been bypassed.}
\label{fig:robustness_examples}
\end{figure*}

\end{document}